\AtBeginDocument{%
  \paperwidth=\dimexpr
    1in + \oddsidemargin
    + \textwidth
    + 1in + \oddsidemargin
  \relax
  \paperheight=\dimexpr
    1in + \topmargin
    + \headheight + \headsep
    + \textheight
    + 1in + \topmargin
  \relax
  \usepackage[pass]{geometry}\relax
}
\RequirePackage{fix-cm}
\documentclass[smallcondensed]{svjour3}     
\smartqed  
\usepackage{amssymb,amsmath,epsfig, booktabs,MnSymbol,commath,subfigure,longtable}
\usepackage{psfrag,graphicx,url,array, afterpage,adjustbox, wrapfig,makecell,mathtools,floatrow}
\usepackage{epstopdf}
\usepackage{hyperref}
\usepackage{comment, natbib, multirow}
\usepackage{xcolor}

\usepackage{epsfig}

\newcolumntype{L}[1]{>{\raggedright\let\newline\\\arraybackslash\hspace{0pt}}m{#1}}
\newcolumntype{C}[1]{>{\centering\let\newline\\\arraybackslash\hspace{0pt}}m{#1}}
\newcolumntype{R}[1]{>{\raggedleft\let\newline\\\arraybackslash\hspace{0pt}}m{#1}}
\newcommand{\x}{\boldsymbol{x}}
\newcommand{\y}{\boldsymbol{y}}
\newcommand{\z}{\boldsymbol{z}}

\newcommand{\R}{\mathbb{R}}
\newcommand{\cmmnt}[1]{\ignorespaces}

\begin{document}

\title{Generative Adversarial Networks in Human Emotion Synthesis:A Review
}
\subtitle{}


\author{Noushin Hajarolasvadi         \and
        Miguel Arjona Ramírez         \and
        Hasan Demirel 
}


\institute{N. Hajarolasvadi \at
              Eastern Mediterranean University, Electrical and Electronic Engineering Department, Gazimagusa, 10 Via Mersin, Turkey \\
              Tel.: +392-630-1384\\
              \email{noushin.hajarolasvadi@cc.emu.edu.tr}           
           \and
           M. A. Ramírez \at
           University of São Paulo, Escola Politécnica, São Paulo, Brazil
           \and
           H. Demirel \at
              Eastern Mediterranean University, Electrical and Electronic Engineering Department, Gazimagusa, 10 Via Mersin, Turkey
}


\maketitle

\begin{abstract}
Synthesizing realistic data samples is of great value for both academic and industrial communities. Deep generative models have become an emerging topic in various research areas like computer vision and signal processing. Affective computing, a topic of a broad interest in computer vision society, has been no exception and has benefited from generative models. In fact, affective computing observed a rapid derivation of generative models during the last two decades. Applications of such models include but are not limited to emotion recognition and classification, unimodal emotion synthesis, and cross-modal emotion synthesis. As a result, we conducted a review of recent advances in human emotion synthesis by studying available databases, advantages, and disadvantages of the generative models along with the related training strategies considering two principal human communication modalities, namely audio and video. In this context, facial expression synthesis, speech emotion synthesis, and the audio-visual (cross-modal) emotion synthesis is reviewed extensively under different application scenarios. Gradually, we discuss open research problems to push the boundaries of this research area for future works.

\end{abstract}

\section{Introduction}\label{intro}
Deep learning techniques are known best for their promising success in uncovering the underlying probability distributions over various data types in the field of artificial intelligence. Some of these data types are videos, images, audio samples, biological signals, and natural language corpora. The success of the deep discriminative models owes primarily to the back-propagation algorithm and piece-wise linear units  \citep{lecun1998gradient, krizhevsky2012imagenet}. In contrast, deep generative models \citep{goodfellow2014generative} have been less successful in addressing deep learning due to difficulties that arise by intractable approximation in the probabilistic computation of methods like maximum likelihood estimation.

Many reviews studied the rapidly expanding topic of generative models and specifically Generative Adversarial Networks (GAN) by investigating various points of view. From algorithms, theory, and applications \citep{gui2020review,wang2017generative} and recent advances and developments  \citep{pan2019recent,zamorski2019generative} to  comparative studies\citep{hitawala2018comparative}, GAN taxonomies \citep{Wang2019GenerativeAN}, and its variants   \citep{hong2019generative,creswell2018generative,huang2018introduction,kurach2018gan} are investigated by the researchers. Also, few review papers discussed the subject based on a specific application like medical imaging \citep{yi2019generative}, audio enhancement and synthesis \citep{torres975audio}, image synthesis \citep{wu2017survey}, and text synthesis\citep{agnese2019survey}. Howsoever, none of the existing surveys considered GAN in view of human emotion synthesis.

\begin{figure*}[ht]
  \centering
  \subfigure[WOS repository]{\includegraphics[width=0.48\textwidth]{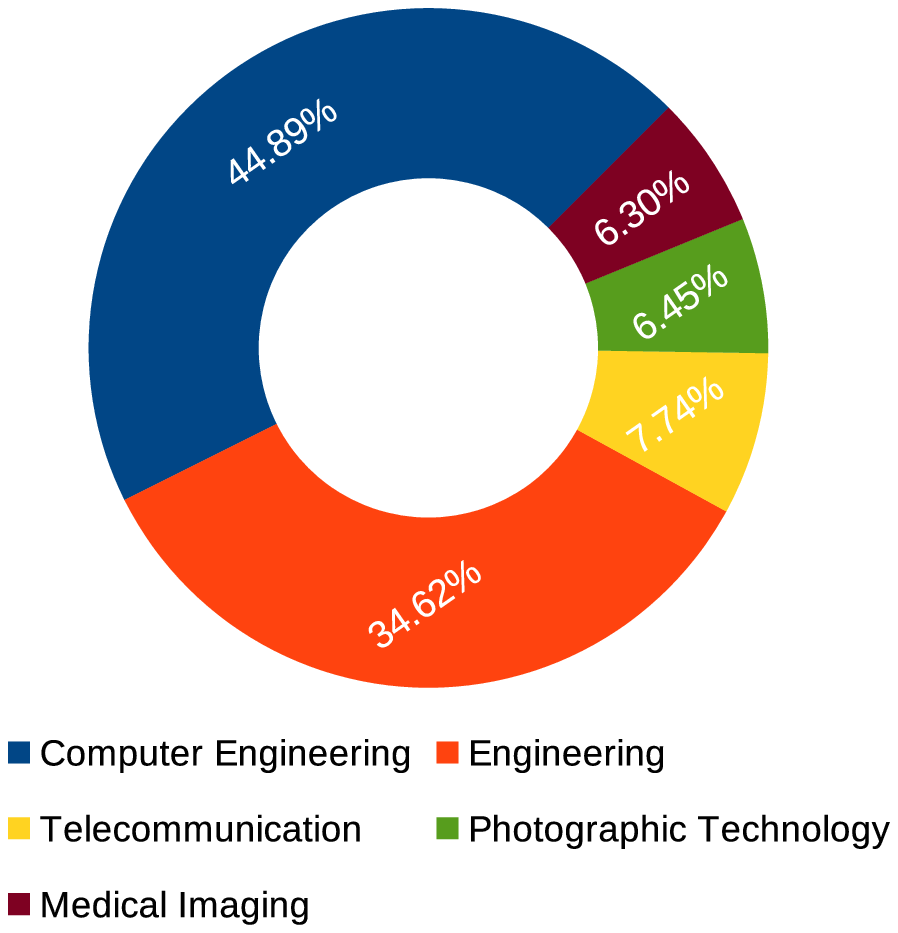}\label{subfig:WOSReport}}\quad
  \subfigure[Scopus repository]{\includegraphics[width=0.48\textwidth]{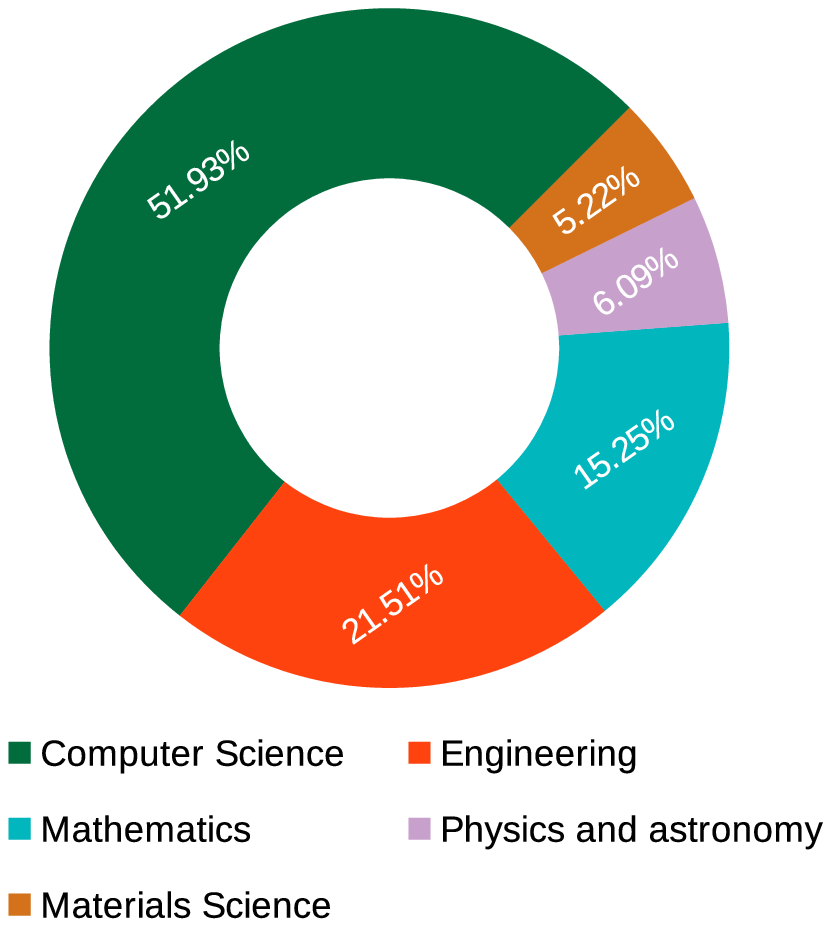}\label{subfig:ScopusReport}}
  
  \caption{Percentage of publications on GAN models since 2014 based on various research areas and by searching the scholarly literature of "Generative Adversarial Networks" studies.}\label{fig:Report}
\end{figure*}

It is important to note that searching the phrase "Generative Adversarial Network" on Web Of Science (WOS) and Scopus repositories report that 2538 and 4405 documents are published, respectively starting from 2014 up to present. Figure \ref{subfig:WOSReport} and \ref{subfig:ScopusReport} show the statistical results obtained from these repositories by searching the aforementioned phrase. The large number of researches published on this topic within only 6 years inspired us to conduct a comprehensive review considering one of the significant applications of GAN models called human emotion synthesis.

Synthesizing realistic data samples is of great value for both academic and industrial communities. Affective computing, a topic of a broad interest in computer vision society benefits from human emotion synthesis and data augmentation. Throughout this paper, we concentrate on the recent advances in the field of GAN and their possible acquisition in the field of human emotion recognition which is known to be useful in other research areas like computer-aided diagnosis systems, security and identity verification, multimedia tagging systems, and human-computer and human-robot interactions. Humans communicate through various verbal and nonverbal channels to show their emotional state. All of the communication modalities are of high importance once interpreting the current emotional state of the user. In this paper, we focus on the GAN-related works of speech emotion synthesis, face emotion synthesis, and audio-visual (cross-modal) emotion synthesis because face and speech are known as pioneer communication channels among humans \citep{schirmer2017emotion, ekman1988smiles, zuckerman1981controlling, mehrabian1967inference}. Researchers developed many GAN-based models to address problems such as data augmentation, improvement of emotion recognition rate, and enhancement of synthesized samples through unimodal \citep{ding2018exprgan},\citep{choi2018stargan,tulyakov2018mocogan},\citep{kervadec2018cake},\citep{kim2018deep},\citep{pascual2017segan},\citep{latif2017variational},\citep{gideon2019improving},\citep{zhou2017mojitalk},\citep{wang2018sentigan} and cross-modal analysis \citep{duarte2019wav2pix},\citep{karras2017audio},\citep{jamaludin2019you},\citep{chen2017deep}.

A specific type of neural network called GAN models was introduced in 2014 by Goodfellow et al. \citep{goodfellow2014generative}. This model is composed of a generative model pitting against an adversary model as a two-player minimax framework. The generative model captures data distribution. Then, given a sample, the adversary or the discriminator decides if the sample is drawn from the true data distribution (real) or the model distribution (fake). The competition continues until the generated samples are indistinguishable from the genuine ones. 

This review deals with the GAN-based algorithms, theory, and applications in human emotion synthesis and recognition. The remainder of the paper is organized as follows: Section \ref{sec:bckg} provides a brief introduction to GANs and their variations. This is followed by a comprehensive review of related works on human emotion synthesis tasks using GANs in section \ref{sec:apps}. This section covers unimodal and cross-modal GAN-based methods developed using audio/visual modalities. Finally, section \ref{sec:disc} summarizes the review, identifies potential applications, and discusses challenges. 

\section{Background} \label{sec:bckg}
In general, generative models can be categorized into explicit density models and implicit density models. While the former utilizes the true data distribution or its parameters to train the generative model, the latter generates sample instances without an explicit parameter assumption or direct estimation on real data distribution. Examples of explicit density modeling are maximum likelihood estimation and Markov Chain Method \citep{kingma2013auto, rezende2014stochastic}. GANs can be considered as implicit density modeling example \citep{goodfellow2014generative}. 

\subsection{Generative Adversarial Networks (GAN)}
\citeauthor{goodfellow2014generative} proposed Generative Adversarial Networks or vanilla GAN in \citeyear{goodfellow2014generative} \citep{goodfellow2014generative}. The model works based on a two-player minimax game where one player seeks to maximize a value function and the other seeks to minimize it. The game ends at a saddle point when the first agent and the second agent reach a minimum and a maximum, respectively, concerning their strategies. This model draws samples directly from the desired distribution without explicitly modeling the underlying probability density function. The general framework of this model consists of two neural network components: a generative model $G$ capturing the data distribution and a discriminative model $D$ estimating the probability that a sample comes from the training samples or $G$. 

\begin{figure}[h]
  \subfigure[vanilla GAN]{\includegraphics[width=.47\textwidth]{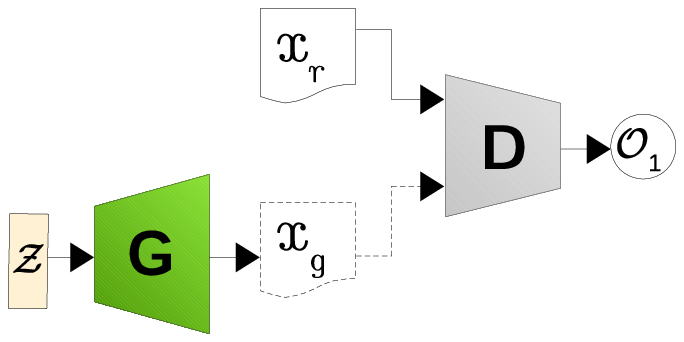}\label{subfig:ganfrmwrk}}\qquad
  \subfigure[CGAN]{\includegraphics[width=0.47\textwidth]{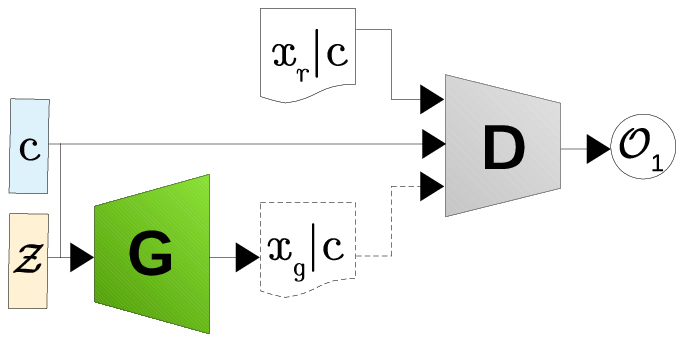}\label{subfig:cganfrmwrk}}\quad
   
  \caption{General Structure of vanilla GAN and CGAN models; $\mathcal{Z}$: input noise, \textbf{G}: \textbf{G}enerator, \textbf{D}: \textbf{D}iscriminator, $\x_r$: \textbf{r}eal sample, $\x_g$: \textbf{g}enerated sample, $\mathcal{O}_1$: \textbf{O}utput of binary classification to real/fake, $c$: \textbf{c}ondition vector.}\label{fig:ganmodels}
\end{figure}

Let us designate the input sample for $G$ as $\z$ where $\z$ is a random noise vector sampled from a priori distribution $p_z(\z)$. Let us denote a real sample as $x_r$ that is taken from the data distribution $P_r$. Also, we show an output sample generated by $G$ as $x_g$. Then, the idea is to get maximum visual similarity between the two samples. In fact, the generator $G$ learns a nonlinear mapping function parametrized by $\theta_g$ and formulated as: $G(\z;\theta_g)$. The discriminator $D$, gets both $x_r$ and $x_g$ to output a single scalar value $\mathcal{O}_1 = D(\x;\theta_d)$ stating the probability that whether an input sample is a real or a generated sample \citep{goodfellow2014generative}. It is important to highlight that $D(\x;\theta_d)$ is the mapping function learned by $D$ and parametrized by $\theta_d$. The final distribution formed by generated samples is $P_g$ and it is expected to approximate $P_r$ after learning. Figure \ref{subfig:ganfrmwrk} illustrates the general block diagram of the vanilla GAN model.

Having two distributions $P_r$ and $P_g$ on the same probability space $\mathcal{X}$, the KL divergence is as follows:

\begin{equation}
    \mbox{KL}(P_r \mid\mid P_g) = \int_{\mathcal{X}} \log\Big(\frac{dP_r(x)}{dP_g(x)} \Big) dP_r(x)
    \label{kld}
\end{equation}

where both $P_r$ and $P_g$ are assumed to admit densities with respect to a common measure defined on $\mathcal{X}$. This happens when $P_r$ and $P_g$ are absolutely continuous, that is $P_g \ll P_r$. The KL divergence is asymmetric, i.e $\mbox{KL}(P_r \mid\mid P_g) \neq \mbox{KL}(P_g \mid\mid P_r)$ and possibly infinite when there are points such that $p_g(x) =0$ and $p_r(x) > 0$ for $KL(P_r \mid\mid P_g)$. A more convenient approach for GAN is the Jensen-Shannon (JS) divergence which may interpreted as a symmetrical version of KL divergence and it is defined as follows:
\begin{equation}
    \mbox{JS}(P_r,P_g)= \mbox{KL}(P_r \mid\mid P_m) + \mbox{KL}(P_g \mid\mid P_m),
    \label{eq:jsd}
\end{equation}
where $ P_m = (P_r + P_g)/2$. 

In other words, a minimax game between $G$ and $D$ continues to obtain a normalized and symmetrical score in terms of the value function $V(G, D)$ as follows: 
\begin{equation}
    \min_{G} \max_{D} V(G,D)=\mathbb{E}_{\x_r \sim p_{r}(\x)}[\log D(\x_r)]+\mathbb{E}_{\z \sim p_{z}(\z)}[\log(1-D(G(\z)))]
    \label{eq:gan}
\end{equation}

Here, the parameters of $G$ are adjusted by minimizing $\log(1-D(G(\x_g)))$. In a similar way, adjusting the parameters for $D$ is performed by maximizing $\log D(\x_r)$. Minimizing $\log(1-D(G(\x_g)))$ is known \citep{goodfellow2014generative} to be equivalent to minimizing the JS divergence between $P_r$ and $P_g$ as expressed in Eq. (\ref{eq:jsd}). The value function $V(\theta_g, \theta_d)$ determines the payoff of the discriminator. Also, the generator takes the value $-V(\theta_g, \theta_d)$ as its own payoff. The generator and the discriminator, each attempts to maximize its own payoff \citep{goodfellow2016deep} during the learning process.The general framework of this model is shown in Figure \ref{subfig:ganfrmwrk}.

\subsection{Challenges of GAN Models}
The training objective of GAN models is often referred to as saddle point optimization problem \citep{yadav2017stabilizing} which is resolved by gradient-based methods. One challenge here is that $D$ and $G$ should be trained at a time so that they advance and converge together. Minimizing the generators' objective is proven to be equivalent to minimizing JS divergence if the discriminator$D$ is trained to its optimal point before the next update of $G$. This means minimizing the JS divergence does not guarantee finding the equilibrium point between $G$ and $D$ through the training process. This normally leads to a better performance of $D$ as opposed to $G$. Consequently, at some point classifying real and fake samples becomes such an easy task that gradients of $D$ approach zero and it becomes ineffectual in the learning procedure of $G$. 
Mode collapse is another well-known problem in training GAN models where $G$ produces a limited set of repetitive samples due to focusing on a few limited modes of the true data distribution, namely $P_r$ during learning and approximating distribution $P_g$. We discuss these problems in more detail in section \ref{ssec:discdis}.

\subsection{Variants by Architectures}\label{subsec:varArch}
The GAN model can be extended to a conditional GAN (CGAN) model if both the generator and discriminator are conditioned on some extra information $\y$ \citep{mirza2014conditional}. Figure \ref{subfig:cganfrmwrk} shows the block diagram of the CGAN model. The condition vector $\y=c$ is fed into both the discriminator and the generator through an additional input layer. Here, the latent variable $\z$ with prior density $p_z(\z)$ and condition vector $\y$ with some value $c \in \R^d$ are passed through one perceptron layer to learn the joint hidden representation. Conditioning on $c$ changes the training criterion of Eq. (\ref{eq:gan}) and leads to the following criterion:

\begin{align}
    \min_{G} \max_{D} V(G,D) =& \mathbb{E}_{\x \sim p_{r}(\x)}[\log D(\x_r\mid \y=c)] \\ + & \nonumber \mathbb{E}_{\z \sim p_{z}(\z)}[\log(1-D(G(\z \mid \y=c)))],
    \label{eq:cgan}
\end{align}
where $c$ could be target class labels or auxiliary information from other modalities. 

\begin{figure}[h]
  \centering
  \includegraphics[width=.9\textwidth]{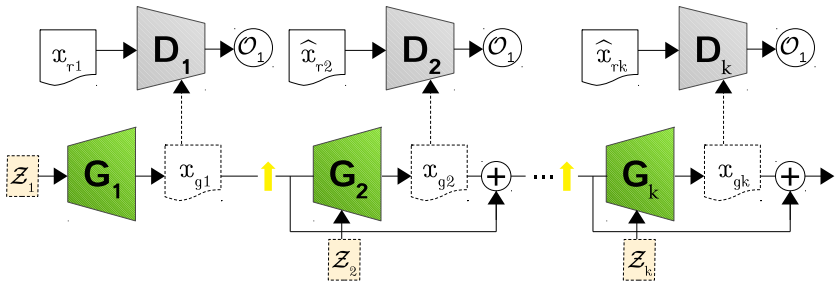}
  \caption{Block diagram of LAPGAN model; \boldmath{$G_k$}: \textit{k}-th \textbf{G}enerator, \boldmath{$D_k$}: \textit{k}-th \textbf{D}iscriminator, $\mathcal{X}_{r1}$: \textbf{r}eal sample, $\mathcal{X}_{rk}$: \textit{k}-th \textbf{r}eal residual, $\mathcal{X}_{gk}$: \textbf{g}enerated sample, $\mathcal{O}_1$: \textbf{O}utput of binary classification to real/fake.}\label{fig:lapganfrmwrk}
\end{figure}

Another type of GAN models is Laplacian Generative Adversarial Network (LAPGAN) \citep{denton2015deep} that formed by combining CGAN models progressively within a Laplacian pyramid representation. LAPGAN includes a set of generative convolutional models, say ${G_1,\dots,G_K}$. The synthesis procedure consists of two parts a sampling phase and a training phase. The sampling phase starts with generator $G_1$ that takes a noise sample $z_1$ and generates sample $x_{g_{1}}$. The generated sample is upscaled before passing to the generator of next level as a conditioning variable. $G_2$ takes both upscaled version of $x_{g_{1}}$ and a noise sample $z_2$ to synthesize a difference sample called $h_2$ which is added to the upscaled version of $x_{g_{1}}$ . This process of upsampling and addition repeats across successive levels to yield a final full resolution sample. The Figure \ref{fig:lapganfrmwrk} illustrates the general block diagram of the LAPGAN model.

SGAN is a second example formed by top-down stacked GAN models \citep{huang2017stacked} to solve the low performance of GAN models in discriminative tasks with large variation in data distribution. \cite{huang2017stacked} employ the hierarchical representation in a model trained discriminatively by stitching GAN models in a top-down framework and forcing the top-most level to take class labels and the bottom-most one to generate images. Alternatively, instead of stacking GANs on top of each other, \citet{karras2017progressive} increased the depth of both the generator and the discriminator by adding new layers. All models are developed under conditional GAN  \citep{denton2015deep, huang2017stacked, karras2017progressive}.

Other models modify the input to the generator slightly. For instance, in SPADE \citep{park2019semantic} a segmentation mask is fed indirectly to the generator through an adaptive normalization layer instead of utilizing the standard input noise vector $\z$. Also, StyleGAN \citep{wang2018high} injects $\z$, first to an intermediate latent space that helps to avoid entanglement of the input latent space to the probability density of the training data.

In \citeyear{radford2015unsupervised}, \citeauthor{radford2015unsupervised} \citep{radford2015unsupervised} proposed Deep Convolutional Generative Adversarial Network (DCGAN) in which both the generator and the discriminator were formed by a class of architecturally constrained convolutional networks. In this model, fully convolutional downsampling/upsampling layers replaced the Fully connected layers of vanilla GAN along with other architectural restrictions like using batch-normalization layers and LeakyReLU activation function in all layers of the discriminator.

Another advancement in GAN models includes using the spectral normalization layer to adjust feature response criterion by normalizing the weights in the discriminator network \citep{miyato2018spectral}. Residual connections are another novel approach fetched into the GAN models by \cite{gulrajani2017improved} and \cite{miyato2018spectral}. While models like CGANs incorporate the conditional information vector simply by concatenation, others remodeled the usage of a conditional vector by a projection approach leading to significant improvement in the quality of the generated samples \citep{miyato2018cgans}.

The aforementioned GAN models expanded based on Convolutional Neural Networks (CNN). Further, along this line, a whole new research line of GAN models developed based on recent deep learning models called CapsuleNets (CapsNets) \citep{sabour2017dynamic}. Let $\boldsymbol{\mathrm{v_k}}$ be the output vector of the final layer of a CapsNet that represents the presence of a visual entity by classifying to one of the $K$ classes. \cite{sabour2017dynamic} provide an updated objective function that benefits from CapsNet margin loss ($L_M$) and it could be expressed as follows:

\begin{equation}\label{eq:marginloss}
    L_M=\sum_{k=1}^{K}T_k\max(0,m^+-\left\|\boldsymbol{\mathrm{v_k}}\right\|)^2+\lambda(1-T_k)\max(0,\left\|\boldsymbol{\mathrm{v_k}}\right\|-m^-)
\end{equation}
where $m^+$, $m^-$, and $\lambda$ are down-weighting factors set to 0.9, 0.1, and 0.5, respectively to stop initial learning from shrinking the lengths of the capsule outputs in the final layer. The length of each capsule in the final layer ($\norm[1]{v_k}$) can then be viewed as the probability of the image belonging to a particular class ($k$). Also, $T_k$ denotes the target label. 

CapsuleGAN \citep{jaiswal2018capsulegan} is a GAN model proposed by \cite{jaiswal2018capsulegan} based on CapsNet.The authors use CapsNet in the discriminator as opposed to conventional CNNs. The final layer of this discriminator consists of a single capsule representing the probability of being a real or fake sample. They used the margin loss introduced in Eq. (\ref{eq:marginloss}) instead of the binary cross-entropy loss for training. The training criterion of the CapsuleGAN is then formulated as follows:

\begin{align}
    \min_{G} \max_{D} V(G,D) =& \mathbb{E}_{\x \sim p_{r}(\x)}[-L_M(D(\x_r), \boldsymbol{T=1})] \\ + &
    \nonumber \mathbb{E}_{\z \sim p_{z}(\z)}[-L_M(D(G(\z)), \boldsymbol{T=0})]
\end{align}

Practically, the generator must be trained to minimize $L_M(D(G(\z)), \boldsymbol{T=1})$ rather than minimizing $-L_M(D(G(\z)), \boldsymbol{T=0})$. This helps eliminating the downweighting factor $\lambda$ in $L_M$ when training the generator, which does not contain
any capsules.

\subsection{Variants by Discriminators}\label{subsec:varDisc}
Stabilizing the training and avoiding mode collapse problem could be achieved by employing different loss functions for $D$. An entropy-based loss is proposed by \cite{springenberg2015unsupervised} called Categorical GAN (CatGAN) in which the objective of discriminator changed from real-fake classification to entropy-based class predictions. WGAN \citep{arjovsky2017wasserstein} and an improved version of it called WGAN-GP \citep{gulrajani2017improved} are two GAN models with a loss function based on Wasserstein distance used in the discriminator. The Earth-Mover (EM) distance or Wasserstein-1 is expressed as follows: 

\begin{equation}\label{wass}
    W(P_r,P_g)= \inf_{\gamma \in \prod(P_r,P_g)} \mathbb{E}_{(\x,\y)\sim\gamma}[\norm[1]{\x-\y}],
\end{equation}
where $\prod(P_r,P_g)$ is the set of all joint distributions $\gamma(\x,\y)$ whose marginals are respectively, $P_r$ and $P_g$. Here, $\gamma(\x,\y)$ describes how much “mass'' needs to be transported from $\x$ to $\y$ in order to transform the distribution $P_r$ into $P_g$. The EM distance is then the “cost" of the optimal transport plan.

\begin{figure}[h]
  \subfigure[CatGAN]{\includegraphics[width=.43\textwidth]{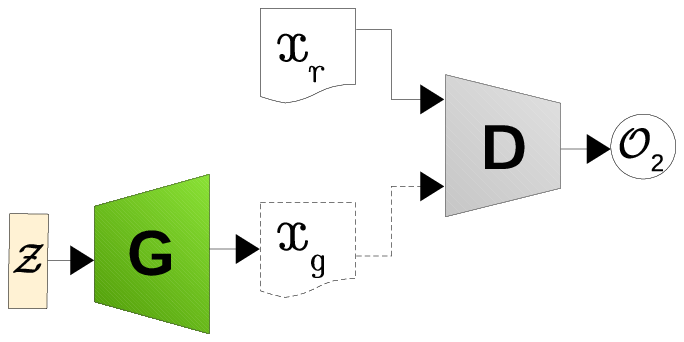}\label{subfig:catganfrmwrk}}\qquad
  \subfigure[BEGAN/EBGAN]{\includegraphics[width=0.43\textwidth]{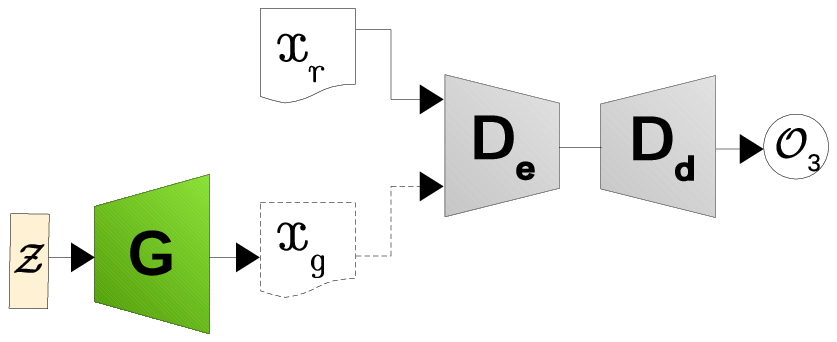}\label{subfig:beganfrmwrk}}
  
  \subfigure[InfoGAN]{\includegraphics[width=.43\textwidth]{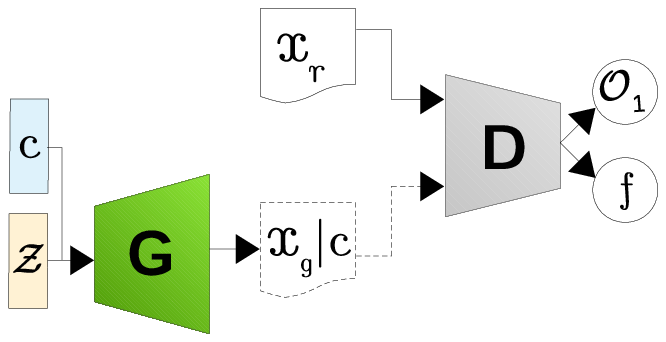}\label{subfig:infoganfrmwrk}}\qquad
  \subfigure[ACGAN]{\includegraphics[width=0.43\textwidth]{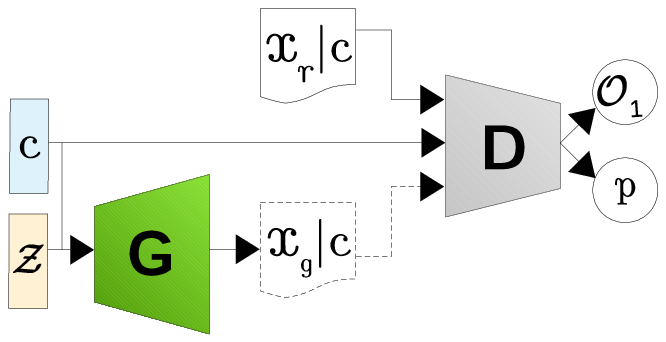}\label{subfig:acganfrmwrk}}
  
  \caption{General Structure of GAN models varying by discriminator; $\mathcal{Z}$: input noise, \textbf{G}: \textbf{G}enerator, \textbf{D}: \textbf{D}iscriminator, $\x_r$: \textbf{r}eal sample, $\x_g$: \textbf{g}enerated sample, $\mathcal{O}_1$: \textbf{O}utput of binary classification to real/fake, $\mathcal{O}_2$: \textbf{O}utput of category classification, $\mathcal{O}_3$: \textbf{O}utput of reconstruction loss for binary classification to real/fake, $c$: condition vector, \boldmath{$D_e$}: incorporated \textbf{D}iscriminator-\textbf{e}ncoder, \boldmath{$D_d$}: incorporated \textbf{D}iscriminator-\textbf{d}ecoder, $f$: a semantic feature vector extracted from $x_g$, $p$: confidence level.}\label{fig:gand}
\end{figure}

Other alternative models that benefit from a different loss metric are GAN based on Category Information (CIGAN) \citep{niu2018cigan}, hinge loss\citep{miyato2018spectral}, least-square GAN \citep{mao2017least}, and f-divergence GAN \citep{nowozin2016f}. Research developments include replacing the encoder structure of the discriminator with an autoencoder structure. In fact, a new loss objective is defined for the discriminator which corresponds to the autoencoder loss distribution instead of data distribution. Examples of such GAN frameworks are Energy-based GAN (EBGAN) \citep{zhao2016energy} and Boundary Equilibrium GAN (BEGAN) \cite{berthelot2017began}. Figure \ref{fig:gand} illustrates the block diagram of GAN models developed by modification in the discriminator.

Another interesting GAN model proposed by \cite{chen2016infogan} is Information Maximizing Generative Adversarial Net (InfoGAN), which simply modifies the discriminator to output both the fake/real classification result and the semantic features of $x_g$ illustrated as $f$ in Figure \ref{subfig:infoganfrmwrk}. The discriminator performs real/fake prediction by maximizing the mutual information between the $x_g$ and conditional vector $c$. Other models like CIGAN \citep{niu2018cigan} and ACGAN \citep{odena2017conditional} focused on improving the quality of the generated samples by employing the class labels during synthesis and then impelling $D$ to provide entropy loss information as well as class probabilities. The Figure \ref{subfig:acganfrmwrk} shows the structure of ACGAN.

\subsection{Variants by Generators}\label{subsec:varGen}
\begin{figure}[h]
    \subfigure[BiGAN/ALI]{\includegraphics[width=.43\textwidth]{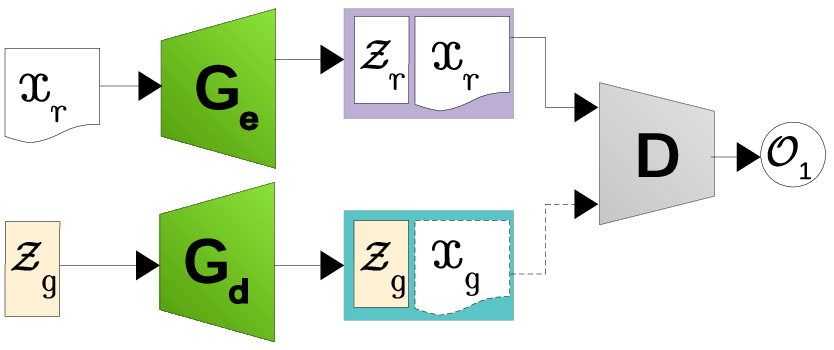}\label{subfig:biganfrmwrk}}\quad
    \subfigure[VAEGAN]{\includegraphics[width=.53\textwidth]{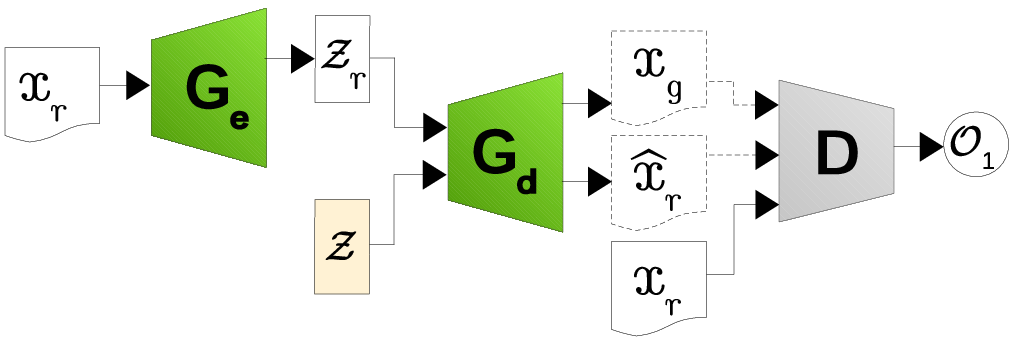}\label{subfig:vaeganfrmwrk}}
  
    \caption{General Structure of UNIT GAN model; $\mathcal{Z}$: input noise, $A$: domain A, $B$: domain B, $\x_r^A$ and $\x_r^B$: \textbf{r}eal sample taken from domain A and B, respectively, $\x_g^A$ and $\x_g^B$: \textbf{g}enerated sample from domain A or B, respectively, $\hat{\x}_r^A$ and $\hat{\x}_r^B$: \textbf{g}enerated fake sample from domain A or B, respectively, $\mathcal{O}_1$: \textbf{O}utput of binary classification to real/fake, \boldmath{$G_e$}: incorporated \textbf{G}enerator-\textbf{e}ncoder, \boldmath{$G_d$}: incorporated \textbf{G}enerator-\textbf{d}ecoder, $\mathcal{Z}_r$: latent vector that encodes the input for $G_e$.}\label{fig:gang}
\end{figure}

The objective of generators is to transform noise input vector $\z$ to a sample $x_g=G(\z)$. In the standard vanilla GAN, this objective is achieved by successively improving the state of the generated sample. The procedure stops when the desired quality is captured. Variational AutoEncoder GAN network (VAEGAN) \citep{larsen2015autoencoding} is arguably the most popular GAN model proposed by varying on the generator architecture. The VAEGAN computes the reconstruction loss in a pixel-wise approach. The decoder network of VAE outputs patterns resembling the true samples (see Figure \ref{subfig:vaeganfrmwrk}).

One challenge in designing GAN models is controlling the attributes of the generated data known as a mode of data. Using supplemental information leads to sample generation with control over the modification of the selected properties. The generator output then becomes $x_g=G(\z,c)$. GANs lack the capability of interpreting the underlying latent space that encodes the input sample. ALI \citep{dumoulin2016adversarially} and BiGAN \citep{donahue2016adversarial} are proposed to resolve this problem by embedding an encoder network in the generator as shown in Figure \ref{subfig:biganfrmwrk}. Here, the discriminator performs real/fake prediction by distinguishing between the tuples $(z_g, x_g)$ and $(z_r, x_r)$. This can categorize the model as a discriminator variant as well.

\begin{figure}[h]
    \subfigure[CycleGAN (unaligned training pairs)]{\includegraphics[width=.45\textwidth]{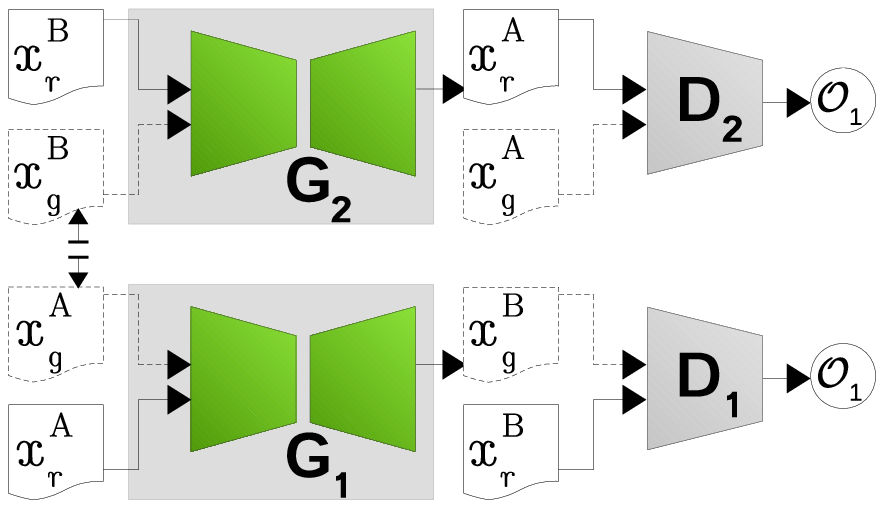}\label{subfig:cycleganfrmwrk}}\qquad    \subfigure[pix2pix (aligned training pairs)]{\includegraphics[width=.45\textwidth]{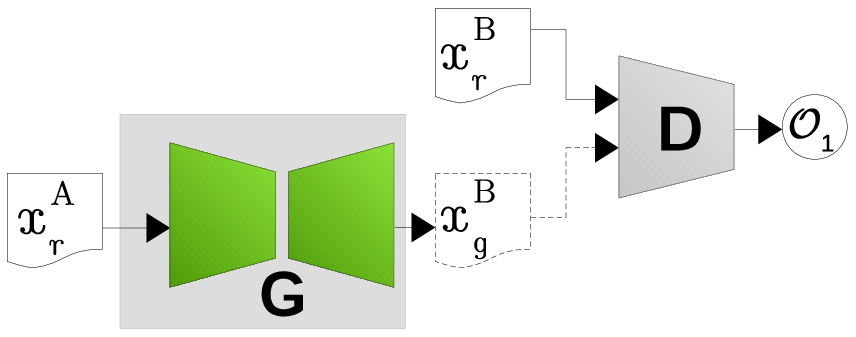}\label{subfig:pix2pixfrmwrk}}
    
    \subfigure[UNIT]{\includegraphics[width=.54\textwidth]{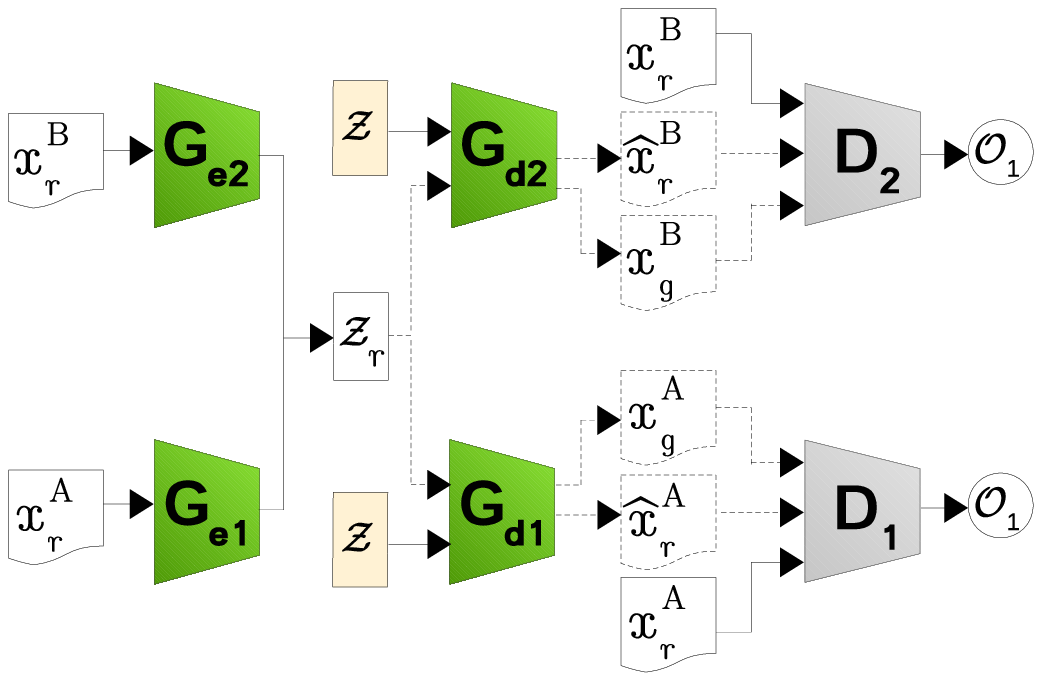}\label{subfig:unitganfrmwrk}}
  
    \caption{General Structure of GAN models varying by generator; $\mathcal{Z}$: input noise, $A$: domain A, $B$: domain B, \textbf{G}:Generator, \textbf{D}:Discriminator, $\x_r^A$ and $\x_r^B$: real sample taken from domain A and B, respectively,  $\x_g^A$ and $\x_g^B$ :generated sample from domain A or B, respectively, $\hat{\x}_r^A$ and $\hat{\x}_r^B$: generated fake sample from domain A or B, respectively, $\mathcal{O}_1$: \textbf{O}utput of binary classification to real/fake,  $G_e$: incorporated \textbf{G}enerator-\textbf{e}ncoder, $G_d$: incorporated \textbf{G}enerator-\textbf{d}ecoder, $\mathcal{Z}_r$: latent vector that encodes the input for $G_e$ }\label{fig:taskgan}
\end{figure}

Other researchers developed the generators to solve specific tasks. \citet{isola2017image} designed pix2pix as an image-to-image translation network to study relations between two visual domains and \citet{milletari2016v} proposed VNet with Dice loss for image segmentation. The disadvantage of such networks was the aligned training with paired samples. In \citeyear{zhu2017unpaired}, \citeauthor{zhu2017unpaired} and \citeauthor{kim2017learning} found a solution to perform unpaired image-to-image translation using cycle consistency loss and cross-domain relations, respectively. Here, the idea was to join two generators together to perform translation between sets of unpaired samples. Below, Figures \ref{subfig:cycleganfrmwrk} and \ref{subfig:pix2pixfrmwrk} show block diagrams of the CycleGAN and pix2pix, respectively.

CycleGAN\citep{zhu2017unpaired} and UNIT are successful examples derived from VAEGAN model. Figure \ref{subfig:unitganfrmwrk} illustrates the layout for UNIT framework. It is important to highlight that considering the generators, the conditional input may vary from class labels \citep{mirza2014conditional} and text descriptions \citep{reed2016generative}, \citep{xu2018attngan} to object location and encoded audio features or cross-modal correlations.

\section{Applications in view of Human Emotion}\label{sec:apps}
In this section, we discuss applications of GAN models in human emotion synthesis. We categorize related works into unimodal and cross-modal researches based on audio and video modalities to help the reader discover applications of interest without difficulty. Also, we explain each method in terms of the proposed algorithm and its advantages and disadvantages. Generally, applications of GAN for human emotion synthesis focus on two issues. The first one is data augmentation that helps obviating the need for the tedious job of collecting and labeling large scale databases and the second is improving the performance on emotion recognition.

\subsection{Facial Expression Synthesis}\label{subsec:fes}
Facial expression synthesis using conventional methods confronts several important problems. First, most methods require paired training data, and second, the generated faces are of low resolution. Moreover, the diversity of the generated faces is limited. The works reviewed in this section are taken from the computer-vision-related researches that focus on facial expression synthesis.

One of the foremost works on facial expression synthesis was the study by \cite{susskind2008generating} that could embed constraints like "raised eyebrows" on generated samples. The authors build their framework upon a Deep Belief Network (DBN) that starts with two hidden layers of 500 units. The output of the second hidden layer is concatenated with identity and a vector of the Facial Action Coding System (FACS) \citep{ekman1978facial} to learn a joint model of them through a Restricted Boltzmann Machine (RBM) with 1000 logistic hidden units. The trained DBN model is then used to generate faces with different identities and facial Action Units (AU).

Later, with the advent of GAN models, DyadGAN is designed specifically for face generation and it can generate facial images of an interviewer conditioned on the facial expressions of their dyadic conversation partner. ExprGAN \citep{ding2018exprgan} is another model designed to solve the problems mentioned above. ExprGAN has the ability to control both the target class and the intensity of the generated expression from weak to strong without a need for training data with intensity values. This is achieved by using an expression controller module that encodes complex information like expression intensity to a real-valued vector and by introducing an identity preserving loss function. 

Other proposed methods before ExprGAN had the ability to synthesize facial expressions either by manipulating facial components in the input image \citep{yang2011expression, mohammed2009visio,yeh2016semantic} or by using the target expression as a piece of auxiliary information  \citep{susskind2008generating, reed2014learning,cheung2014discovering}. In \citeyear{zhou2017photorealistic}, \citeauthor{shu2017neural} and \citeauthor{zhou2017photorealistic} proposed two GAN-based models. \citeauthor{shu2017neural} learns a disentangled representation of inherent facial attributes by manipulating facial appearance, and \citeauthor{zhou2017photorealistic} synthesizes facial appearance of unseen subjects using AUs and a conditional adversarial autoencoder. 

\begin{center}
    \fontsize{8}{8}\selectfont
    \begin{longtable}{L{1.6cm}L{1.5cm}L{1.75cm}L{.7cm}L{.7cm}L{.7cm}L{.85cm}L{.8cm}}\\
    \toprule
        author & based on & model & loss & data & M & RS & RM\\
    \toprule \\
    \endfirsthead
    \multicolumn{8}{l}%
    {\tablename\ \thetable\ -- \textit{Continued from previous page}} \\
    \toprule
        author & based on & method & loss & data & M & RS & RM\\
    \toprule \\
    \endhead
    \hline \multicolumn{8}{c}{\textit{Continued on next page}} \\
    \endfoot
    \caption{Comparison of facial expression image synthesis models, description of loss functions (L), metrics (M), databases (D) and purposes (P) used in the reviewed publications are given in Tables \ref{tab:fdatabases}, \ref{tab:flosses}, \ref{tab:fmetrics},  and \ref{tab:fpur}\\ \\
    \footnotesize{
    - M: Metric, RS: Results, RM:Remarks \\
    - *: shows the proposed method by authors, other mentioned methods are implemented by the authors for the sake of comparison \\
    - the result reported for expression classification accuracy (M$v_{2}$) belongs to the synthesized image datasets \\
    -$\dagger$: D$v_{9}$ + D$v_{10}$ is used as the database\\
    -$\ddagger$: D$v_{9}$ + D$v_{11}$ is used as the database\\
    - All papers provide visual representation of the synthesized images (M$v_{7}$)\\
    }} \label{tab:FacialImageGANS}
    
    \endlastfoot
    \citeauthor{huang2017dyadgan} & CGAN & DyadGAN & L$v_{1}$ & D$v_{5}$ & M$v_{7}$ & - & \makecell[l]{P$v_{3}$,\\ P$v_{5}$,\\ P$v_{15}$,\\ P$v_{19}$} \\
        \midrule
        \citeauthor{ding2018exprgan} & CGAN & ExprGAN & \makecell[l]{L$v_{1}$,\\ L$v_{2}$,\\ L$v_{3}$,\\ L$v_{4}$,\\ L$v_{5}$} & D$v_{1}$ & M$v_{2}$ & 
        84.72 & \makecell[l]{P$v_{1}$,\\ P$v_{2}$, \\ P$v_{3}$,\\ P$v_{5}$}  \\
        \midrule
        \multirow{8}{*}{\citeauthor{song2018geometry}}  & \multirow{8}{*}{CGAN} & \multirow{8}{*}{G2GAN} & \multirow{8}{*}{\makecell[l]{L$v_{1}$,\\ L$v_{2}$,\\ L$v_{3}$,\\ L$v_{6}$,\\ L$v_{12}$}}  & \multirow{4}{*}{D$v_{1}$} & M$v_{2}$ & 58.94 & \multirow{8}{*}{\makecell[l]{P$v_{1}$,\\ P$v_{2}$,\\ P$v_{5}$,\\ P$v_{10}$ }} \\
        & & & & & M$v_{4}$ & 100 & \\
        & & & & & M$v_{5}$ & 0.94 & \\
        & & &  & & M$v_{6}$ & 29.50 & \\
        \cmidrule{5-7}
        & & & & \multirow{2}{*}{D$v_{2}$} & M$v_{5}$ & 0.85 & \\
        & & & & & M$v_{6}$ & 24.81 & \\
        \cmidrule{5-7}
        & & & & \multirow{2}{*}{D$v_{3}$} & M$v_{5}$ & 0.82 & \\
        & & & & & M$v_{6}$ & 27.30 & \\
        \midrule
        \multirow{3}{*}{\citeauthor{choi2018stargan}} & \multirow{3}{*}{CycleGAN} & \multirow{3}{*}{StarGAN} & \multirow{3}{*}{\makecell[l]{L$v_{1}$,\\ L$v_{6}$,\\ L$v_{7}$}} & \multirow{3}{*}{D$v_{5}$} & M$v_{10}$ & 52.20 & \multirow{3}{*}{\makecell[l]{P$v_{4}$,\\ P$v_{7}$,\\ P$v_{14}$}} \\
        & & & & & & & \\
        & & & & & M$v_{3}$ & 2.12 & \\
        \midrule
        \citeauthor{vielzeuf2019many} & StarGAN & - & \makecell[l]{L$v_{1}$,\\ L$v_{7}$,\\ L$v_{10}$} & D$v_{8}$ & M$v_{3}$ & 3.4 & \makecell[l]{P$v_{2}$,\\ P$v_{9}$,\\ P$v_{11}$} \\
        \midrule
        \multirow{5}{*}{\citeauthor{lai2018emotion}} & \multirow{5}{*}{VAEGAN} & \multirow{5}{*}{-} &\multirow{5}{*}{\makecell[l]{L$v_{1}$,\\ L$v_{2}$,\\ L$v_{7}$,\\ L$v_{8}$,\\ L$v_{13}$}} & D$v_{2}$ & \multirow{5}{*}{M$v_{2}$} & 87.08 & \multirow{5}{*}{\makecell[l]{P$v_{1}$,\\ P$v_{12}$,\\ P$v_{13}$ }} \\
        & & & & & & & \\
        & & & & & & & \\
        & & & & D4 & & 73.13 & \\
        & & & & & & & \\
        \midrule
        \multirow{5}{*}{\makecell[l]{Caramihale\\ et al.}}&\multirow{5}{*}{CycleGAN} & \multirow{5}{*}{-} & \multirow{5}{*}{\makecell[l]{L$v_{1}$,\\ L$v_{6}$}} & D$v_{3}$ & \multirow{5}{*}{M$v_{2}$} & 98.30 & \multirow{5}{*}{\makecell[l]{P$v_{1}$,\\ P$v_{13}$}} \\
        & & & & D$v_{9}$ & & 75.20 & \\
        & & & & D$v_{10}$ & & 60.80 & \\
        & & & & D$v_{11}$ & & 94.80 & \\
        & & & & D$v_{12}$ & & 75.70 & \\
        \midrule
        \multirow{3}{*}{\citeauthor{zhu2018emotion}} & \multirow{3}{*}{CycleGAN} & \multirow{3}{*}{-} & \multirow{3}{*}{\makecell[l]{L$v_{1}$,\\ L$v_{6}$}} & D$v_{9}$ & \multirow{3}{*}{M$v_{2}$} & 94.71 & \multirow{3}{*}{P$v_{1}$} \\
        & & & & $\dagger$ & & 39.07 & \\
        & & & & $\ddagger$ & & 95.80 & \\
        \midrule
        \citeauthor{lindt2019facial} & VAEGAN & - &  \makecell[l]{L$v_{1}$,\\ L$v_{3}$,\\ L$v_{4}$,\\ L$v_{10}$} & D$v_{8}$ & M$v_{3}$ & 6.07 & \makecell[l]{P$v_{3}$,\\ P$v_{4}$,\\ P$v_{9}$} \\
        \midrule
        \citeauthor{lu2018attribute} & CycleGAN & AttGGAN & \makecell[l]{L$v_{1}$,\\ L$v_{3}$,\\ L$v_{6}$} & D$v_{6}$ & M$v_{5}$ & 0.92 & \makecell[l]{P$v_{8}$,\\ P$v_{10}$,\\ P$v_{14}$} \\
        \midrule
        \citeauthor{zhang2018tv} & CGAN & FaceID & \makecell[l]{L$v_{1}$,\\ L$v_{3}$,\\ L$v_{18}$} & D$v_{12}$ & M$v_{4}$ & 97.01 & \makecell[l]{P$v_{2}$,\\ P$v_{4}$,\\ P$v_{13}$} \\
        \midrule
        \multirow{3}{*}{\makecell[l]{Peng and\\ Yin}} & \multirow{3}{*}{CycleGAN} & \multirow{3}{*}{ApprGAN} & \multirow{3}{*}{\makecell[l]{L$v_{1}$,\\ L$v_{3}$,\\ L$v_{10}$}} & D$v_{3}$ & \multirow{3}{*}{M$v_{15}$} & 0.95 & \multirow{3}{*}{P$v_{8}$} \\
        & & & & & & & \\
        & & & & D$v_{14}$ & & 0.97 & \\
        \midrule
        \citeauthor{shen2018faceid} & Pix2Pix & TVGAN & L$v_{1}$, L$v_{3}$ & D$v_{16}$ & M$v_{4}$ & 50.90 & \makecell[l]{P$v_{1}$,\\ P$v_{2}$,\\ P$v_{20}$} \\
        \midrule
        \multirow{7}{*}{\citeauthor{he2019attgan}} & \multirow{7}{*}{\makecell[l]{VAEGAN \\ IcGAN}} & AttGAN* & \multirow{7}{*}{\makecell[l]{L$v_{1}$, \\ L$v_{7}$,\\ L$v_{10}$}} & \multirow{4}{*}{D$v_{6}$} & \multirow{7}{*}{M$v_{11}$} & 88.20 & \multirow{7}{*}{\makecell[l]{P$v_{3}$,\\ P$v_{4}$,\\ P$v_{7}$,\\ P$v_{14}$}}\\
        & & CycleGAN & & & & 67.80 & \\
        & & VAEGAN & & & & 52.00 & \\
        & & StarGAN & & & & 86.20 & \\
        \cmidrule{5-5}\cmidrule{7-7}
        & & AttGAN* & &\multirow{3}{*}{D$v_{12}$} & & 88.00 & \\
        & & CycleGAN & & & & 63.40 & \\
        & & StarGAN &  & & & 73.40 & \\
        \midrule
        & & & L$v_{1}$, & & & & \\
        \multirow{5}{*}{\citeauthor{wang2020learning}} & \multirow{5}{*}{GAN} & CompGAN* & \multirow{5}{*}{\makecell[l]{L$v_{3}$,\\ L$v_{6}$,\\ L$v_{7}$,\\ L$v_{10}$,\\ L$v_{14}$}}& \multirow{5}{*}{D$v_{14}$} &\multirow{5}{*}{M$v_{2}$}& 74.92 & \multirow{5}{*}{\makecell[l]{P$v_{2}$,\\ P$v_{4}$,\\ P$v_{13}$}} \\
        &  & CycleGAN & & & & 65.38 & \\
        &  & pix2pix & & & & 71.31 & \\
        &  & VAEGAN & & & & 70.88 & \\
        &  & AttGAN & & & & 71.92 & \\
        \midrule
        \multirow{8}{*}{\makecell[l]{Borogtabar\\ et al.}} & \multirow{8}{*}{CGAN} & ExprADA* & \multirow{8}{*}{\makecell[l]{L$v_{1}$,\\ L$v_{3}$,\\ L$v_{7}$,\\ L$v_{15}$}}&  \multirow{2}{*}{D$v_{4}$} & \multirow{8}{*}{M$v_{2}$} & 73.20 & \multirow{8}{*}{\makecell[l]{P$v_{1}$,\\ P$v_{5}$,\\ P$v_{7}$,\\ P$v_{13}$,\\ P$v_{14}$}} \\
        & & CycleGAN & & & & 71.60 & \\
        \cmidrule{5-5}\cmidrule{7-7}
        & & ExprADA* & &\multirow{2}{*}{D$v_{17}$} & & 70.7 & \\
        & & CycleGAN & & & & 63.50 & \\
        \cmidrule{5-5}\cmidrule{7-7}
        & & ExprADA* & & \multirow{2}{*}{D$v_{18}$} & & 86.90 & \\
        & & CycleGAN & & & & 71.20 & \\
        \cmidrule{5-5}\cmidrule{7-7}
        & & ExprADA* & & \multirow{2}{*}{D$v_{19}$} & & 86.90 & \\
        & & CycleGAN & & & & 82.40 & \\
        \midrule
        \multirow{3}{*}{\citeauthor{wang2019u}} & \multirow{3}{*}{ACGAN} & \multirow{3}{*}{UNet GAN} & \multirow{3}{*}{\makecell[l]{L$v_{1}$,\\ L$v_{7}$,\\ L$v_{16}$}} & D$v_{1}$ & \multirow{3}{*}{M$v_{2}$} & 43.33 & \multirow{3}{*}{\makecell[l]{P$v_{2}$,\\ P$v_{4}$}} \\
        & & & & & & & \\
         & & & & D$v_{5}$ & & 82.59 & \\
        \midrule
        \makecell[l]{\\ \citeauthor{lee2019collagan}} & \makecell[l]{\\ CycleGAN} & \makecell[l]{\\ CollaGAN} & \makecell[l]{L$v_{1}$,\\ L$v_{6}$, \\ L$v_{7}$,\\ L$v_{17}$} & D$v_{2}$ & M$v_{7}$ & - & \makecell[l]{P$v_{7}$,\\ P$v_{10}$,\\ P$v_{16}$,\\ P$v_{17}$,\\ P$v_{18}$} \\
        \midrule
        \makecell[l]{\\ \citeauthor{shen2018facefeat}} & \makecell[l]{\\ StackGAN} & \makecell[l]{\\ FaceFeat} & \makecell[l]{L$v_{1}$,\\ L$v_{3}$,\\ L$v_{8}$,\\ L$v_{20}$} & D$v_{12}$ & M$v_{4}$ & 97.62 & \makecell[l]{P$v_{2}$,\\ P$v_{4}$} \\
        \midrule
        \multirow{3}{*}{\citeauthor{deng2018uv}} & \multirow{3}{*}{GAN} & \multirow{3}{*}{UVGAN} & \multirow{3}{*}{\makecell[l]{L$v_{1}$,\\ L$v_{3}$,\\ L$v_{20}$}} & \multirow{3}{*}{D$v_{20}$} & M$v_{5}$ & 0.89 & \multirow{3}{*}{\makecell[l]{P$v_{2}$,\\ P$v_{4}$,\\ P$v_{13}$}}\\
        & & & & & & & \\
        & & & & & M$v_{6}$ & 25.06 & \\
        \midrule
        \multirow{2}{*}{\citeauthor{li2017generative}} & \multirow{2}{*}{GAN} & \multirow{2}{*}{-} & \multirow{2}{*}{\makecell[l]{L$v_{1}$,\\ L$v_{3}$}} & \multirow{2}{*}{D$v_{6}$} & M$v_{5}$ & 0.84 & \multirow{2}{*}{\makecell[l]{P$v_{2}$,\\ P$v_{4}$}} \\
        & & & &  & M$v_{6}$ & 20.20 & \\
        \midrule
        \multirow{4}{*}{\citeauthor{cheng2019meshgan}} & \multirow{4}{*}{\makecell[l]{BEGAN \\ ChebNet \\ CoMA}} & MeshGAN* & \multirow{4}{*}{\makecell[l]{L$v_{1}$}} & \multirow{2}{*}{D$v_{21}$} & \multirow{2}{*}{M$v_{16}$} & 1.43 & \multirow{4}{*}{\makecell[l]{P$v_{2}$,\\ P$v_{4}$}} \\
         & & CoMA & & & & 1.60 & \\
         \cmidrule{5-7}
         & & MeshGAN* & & \multirow{2}{*}{D$v_{22}$} & \multirow{2}{*}{M$v_{3}$} & 0.85 & \\
         & & CoMA & & & & 1.89 & \\
        \bottomrule
        & & & & & & & \\
\end{longtable}
\end{center}
Table \ref{tab:FacialImageGANS} compares the reviewed publication based on various metrics, databases, loss functions and purposes used by researchers. Following those models and through the many variations of facial expression synthesis proposed by researchers, the GAN-based model proposed by \citet{song2018geometry} was one of the interesting and premier ones, called G2GAN. G2GAN generates photo-realistic and identity-preserving images. Furthermore, it provides fine-grained control over the target expression and facial attributes of the generated images like widening the smile of the subject or narrowing the eyes. The idea here is to feed the face geometry into the generator as a condition vector which guides the expression synthesis procedure. The model benefits from a pair of GANs that while one removes the expression, the other synthesizes it. This leverages on the ability of unpaired training.
\begin{table*}[b]
    \setlength{\aboverulesep}{0pt}
    \setlength{\belowrulesep}{0pt}
    \caption{List of databases used for facial emotion synthesis in the reviewed publications}
    \label{tab:fdatabases}
    \begin{tabular}{lL{5cm}clcl}
        \toprule
          & Database & \#Subjects & \#samples & \#classes\\
         \toprule
         D$v_{1}$ & Oulu-CASIA (\citeauthor{zhao2011facial}) & 80 & 2,880 & 6\textsuperscript{$\star$} \\
         D$v_{2}$ & MULTI-PIE (\citeauthor{gross2010multi}) & 337 & 755,370 & 6\textsuperscript{$\dagger$}\\
         D$v_{3}$ & CK+ (\citeauthor{lucey2010extended}) & 123 & 593 & 6\textsuperscript{$\diamond$} \\
         D$v_{4}$ & BU-3DFE (\citeauthor{yin20063d}) & 100 & 2,500 & 6\textsuperscript{$\bullet$}\\
         D$v_{5}$ & RaFD (\citeauthor{langner2010presentation}) & 67 & 1,068 & 6\textsuperscript{$\diamond$}\\
         D$v_{6}$ & CelebA (\citeauthor{liu2015deep}) & 10,177 & 202,599 & 45\textsuperscript{$+$} \\
         D$v_{7}$ & EmotioNet (\citeauthor{benitez2017emotionet}) & N/A & 1,000,000 & 23\textsuperscript{$\ddagger$}\\
         D$v_{8}$ & AffectNet (\citeauthor{mollahosseini2017affectnet}) & N/A & 450,000 & 6\textsuperscript{$\bullet$}\\
         D$v_{9}$ & FER2013 (\citeauthor{goodfellow2013challenges}) & N/A & 35,887 & 6\textsuperscript{$\bullet$} \\
         D$v_{10}$ & SFEW (\citeauthor{dhall2011static}) & N/A & 1,766 & 6\textsuperscript{$\bullet$} \\
         D$v_{11}$ & JAFFE (\citeauthor{lyons1998japanese}) & 10 & 213  & 6\textsuperscript{$\bullet$}\\
         D$v_{12}$ & LFW (\citeauthor{huang2008labeled})  & 5,749 & 13,233  & -\\
         D$v_{13}$ & F\textsuperscript{2}ED (\citeauthor{wang2020learning}) & 119 & 219,719  & 54\textsuperscript{$\triangleleft$}\\
         D$v_{14}$ & MUG (\citeauthor{aifanti2010mug}) & 52 & 204,242  & 6\textsuperscript{$\triangleright$}\\
         D$v_{15}$ & Dyadic Dataset (\citeauthor{huang2017dyadgan}) & 31 & - & 8\textsuperscript{$\times$}\\
         D$v_{16}$ & IRIS Dataset (\citeauthor{kong2007multiscale})& 29 & 4,228 & 3\textsuperscript{$\oplus$}\\
         D$v_{17}$ & MMI (\citeauthor{pantic2005web})& 31 & 236 & 6\textsuperscript{$\bullet$}\\
         D$v_{18}$ & Driver emotion (\citeauthor{bozorgtabar2020exprada}) & 26 & N/A & 6\textsuperscript{$\bullet$}\\
         D$v_{19}$ & KDEF (\citeauthor{lundqvist1998karolinska}) & 70 & N/A & 6\textsuperscript{$\bullet$}\\
         D$v_{20}$ & UVDB (\citeauthor{deng2018uv}) & 5,793 & 77,302 & -\\
         D$v_{21}$ & 3dMD (\citeauthor{cheng2019meshgan}) & 12,000 & N/A & -\\
         D$v_{22}$ & 4DFAB (\citeauthor{cheng20184dfab}) & 180 & 1,800,000 & 6\textsuperscript{$\bullet$}\\
         \bottomrule
    \end{tabular}
    \begin{flushleft}
        \scriptsize{
            \hspace{1.65em} D stands for Database, A: Audio, V:Visual, A-V:Audio-Visual\\
            \hspace{1.65em} 6\textsuperscript{$\star$}: 6 basic expressions including angry, disgust, fear, happiness, sadness, and surprise\\
            \hspace{1.65em} 6\textsuperscript{$\dagger$}: 6 classes including smile, surprised, squint, disgust, scream, and neutral\\
            \hspace{1.65em} 6\textsuperscript{$\diamond$}: 6\textsuperscript{$\star$} + neutral and contempt\\
            \hspace{1.65em} 6\textsuperscript{$\bullet$}: 6\textsuperscript{$\star$} + neutral \\
            \hspace{1.65em} 23\textsuperscript{$\ddagger$}: 6 basic expressions + 17 compound emotions\\
            \hspace{1.65em} 45\textsuperscript{$+$}: 5 landmark locations, 40 binary attributes\\
            \hspace{1.65em} 54\textsuperscript{$\triangleleft$}: 54 emotion types (categories are not mentioned clearly in the source paper)\\
            \hspace{1.65em} 6\textsuperscript{$\star$} + neutral, also landmark point annotation is provided)\\
            \hspace{1.65em} 6\textsuperscript{$\times$}: Joy, Anger, Surprise, Fear, Contempt, Disgust, Sadness and Neutral\\
            \hspace{1.65em} 3\textsuperscript{$\oplus$}: surprised, laughing, angry
        }
    \end{flushleft}
\end{table*}

StarGAN \citep{choi2018stargan} is the first approach with a scalable solution for multi-domain image-to-image translation using a unified GAN model (i.e only a single generator and discriminator). In this model, a domain is defined as a set of images sharing the same attribute and attributes are the facial features like hair color, gender, and age which can be modified based on the desired value. For example, one can set hair color to be blond or brown and set the gender to be male or female. Likewise, Attribute editing GAN (AttGAN) \citep{he2019attgan} provides a GAN framework that can edit any attribute among a set of attributes for face images by employing adversarial loss, reconstruction loss, and attribute classification constraints. Also, DIAT \citep{li2016deep}, CycleGAN \citep{zhu2017unpaired} and IcGAN \citep{perarnau2016invertible} could be compared as baseline models.

In \citeyear{qiao2018emotional}, G2GAN \citeauthor{song2018geometry} is extended by \citeauthor{qiao2018emotional}. The authors derived a model based on VAEGANs to synthesize facial expressions given a single image and several landmarks through some transferring stages. Different from ExprGAN their model does not require the target class label of the generated image. Also, unlike G2GAN, it does not require the neutral expression of a specific subject as an intermediate level in the facial expression transfer procedure. While G2GAN and its extension focus on geomterical features to guide the expression synthesis procedure, \citet{pumarola2018ganimation} use facial AU as a one-hot vector to perform an unsupervised expression synthesis while smooth transition and unpaired samples are guaranteed. 

Another VAEGAN-based model is the work of \citep{lai2018emotion} where a novel optimization loss called symmetric loss is introduced. Symmetric loss helps preserving the symmetrical property of the face while translating from various head poses to frontal-view of the face. Similar to \citeauthor{lai2018emotion} is the FaceID-GAN \citep{shen2018faceid} where, in addition to the two-players of vanilla GANs and symmetry information, a classifier of face identity is employed as the third player that competes with the generator by distinguishing the identities of the real and synthesized faces.
\begin{table}[h]
    \setlength{\aboverulesep}{0pt}
    \setlength{\belowrulesep}{0pt}
    \caption{list of loss functions used for facial emotion synthesis in the reviewed publications}
    \label{tab:flosses}
    \begin{tabular}{llL{9cm}}
        \toprule
         & Name & Remarks\\
         \toprule
         L$v_{1}$ &  $\mathcal{L}_{adv}$ & adversarial loss presented by the discriminator, (see section \ref{subsec:varDisc})  \\
         L$v_{2}$ & $\mathcal{L}_{pixel}$ & pixel-wise image reconstruction loss\\
         L$v_{3}$ & $\mathcal{L}_{id}$ & identity preserving loss\\
         L$v_{4}$ & $\mathcal{L}_{regu}$ & loss of a regularizer \\
         L$v_{5}$ & $\mathcal{L}_{tv}$ & total variation regularizer loss\\
         L$v_{6}$ & $\mathcal{L}_{cyc}$ & cycle-consistency loss\\
         L$v_{7}$ & $\mathcal{L}_{cls}$ & classification loss (expression)\\
         L$v_{8}$ & $\mathcal{L}_{feat}$ & feature matching loss\\
         L$v_{9}$ & $\mathcal{L}_{contr}$ & contrastive loss \\
         L$v_{10}$ & $\mathcal{L}_{rec}$ & image reconstruction loss \\
         L$v_{11}$ & $\mathcal{L}_{att}$ & attention loss \\
         L$v_{12}$ & $\mathcal{L}_{cond}$ & conditional loss \\
         L$v_{13}$ & $\mathcal{L}_{sym}$ & Symmetry loss to preserve symmetrical property of the face\\
         L$v_{14}$ & $\mathcal{L}_{pose}$ & cross-entropy loss used to ensure correct pose of the face\\
         L$v_{15}$ & $\mathcal{L}_{bi}$ & bidirectional loss to avoid mode collapse\\
         L$v_{16}$ & $\mathcal{L}_{triplet}$ & triplet loss to minimize the similarity between $\x_r$ and $\x_g$ \\
         L$v_{17}$ & $\mathcal{L}_{SSIM}$ & Structural Similarity Index Loss that measures the image quality\\
         L$v_{18}$ & $\mathcal{L}_{P}$ & learn the shape feature by minimizing the weighted distance\\
         L$v_{19}$ & $\mathcal{L}_{recurr}$ & loss for a recurrent temporal predictor to predict future samples\\
         L$v_{20}$ & $\mathcal{L}_{3DMM}$ & 3D Morphable Model loss to ensure correct pose and expression\\
         L$v_{21}$ & $\mathcal{L}_{motion}$ & motion loss consisting of a VAE loss, a video reconstruction loss, and KL-divergence between the prior and posterior motion latent distribution\\
         L$v_{22}$ & $\mathcal{L}_{cnt}$ & contents loss consisting of a reconstruction loss for the current frame and a KL-divergence between the prior and posterior content distribution\\
        \bottomrule
    \end{tabular}
\end{table}

\cite{lai2018emotion} used GAN to perform emotion-preserving
representations. In the proposed approach, the generator can transform the non-frontal facial images into frontal ones while the identity and the emotion expression are preserved. Moreover, a recent publication \citep{vielzeuf2019many} relies on a two-step GAN framework. The first component maps images to a 3D vector space. This vector is issued from a neural network and it represents the corresponding emotion of the image. Then, a second component that is a standard image-to-image translator uses the 3D points obtained in the first step to generate different expressions. The proposed model provides fine-grained control over the synthesized discrete expressions through the continuous vector space representing the arousal, valence, and dominance space.
\begin{table}[h]
    \setlength{\aboverulesep}{0pt}
    \setlength{\belowrulesep}{0pt}
    \caption{List of evaluative metrics used for facial emotion synthesis in the reviewed publications}
    \label{tab:fmetrics}
    \begin{tabular}{llL{5.25cm}}
         \toprule
         & Measurement & Remarks \\
         \toprule
         M$v_{1}$ &  & Ground truth, costly, not scaleable\\
         M$v_{2}$ & expression classification (accuracy) & down stream task\\
         M$v_{3}$ & expression classification (error) & down stream task\\
         M$v_{4}$ & identity classification (accuracy) & down stream task\\
         M$v_{5}$ & Structural Similarity Index Measure \cmmnt{(SSIM)} & measures image quality degradation\\
         M$v_{6}$ & Peak Signal to Noise Ratio (PSNR) & measures quality of representation\\
         M$v_{7}$ & visual representation & down stream task \\
         M$v_{8}$ & real/fake classification (accuracy) & down stream task\\
         M$v_{9}$ & real/fake classification (error) & down stream task\\
         M$v_{10}$ & attribute classification (accuracy) & down stream task\\
         M$v_{11}$ & attribute classification (error) & down stream task\\
         M$v_{12}$ & Average Content Distance (ACD) & content consistency of a generated video \\
         M$v_{13}$ & Motion Control Score (MCS) & capability in motion generation control\\
         M$v_{14}$ & Inception Score (IS) & measures quality and diversity of $P_g(\x)$\\
         M$v_{15}$ & texture similarity score & measuring texture similarity\\
         M$v_{16}$ & identity classification (error) & down stream task\\
         \bottomrule
    \end{tabular}
\end{table}

\begin{figure}[h]
    \subfigure[ExprGAN]{\includegraphics[width=.8\textwidth]{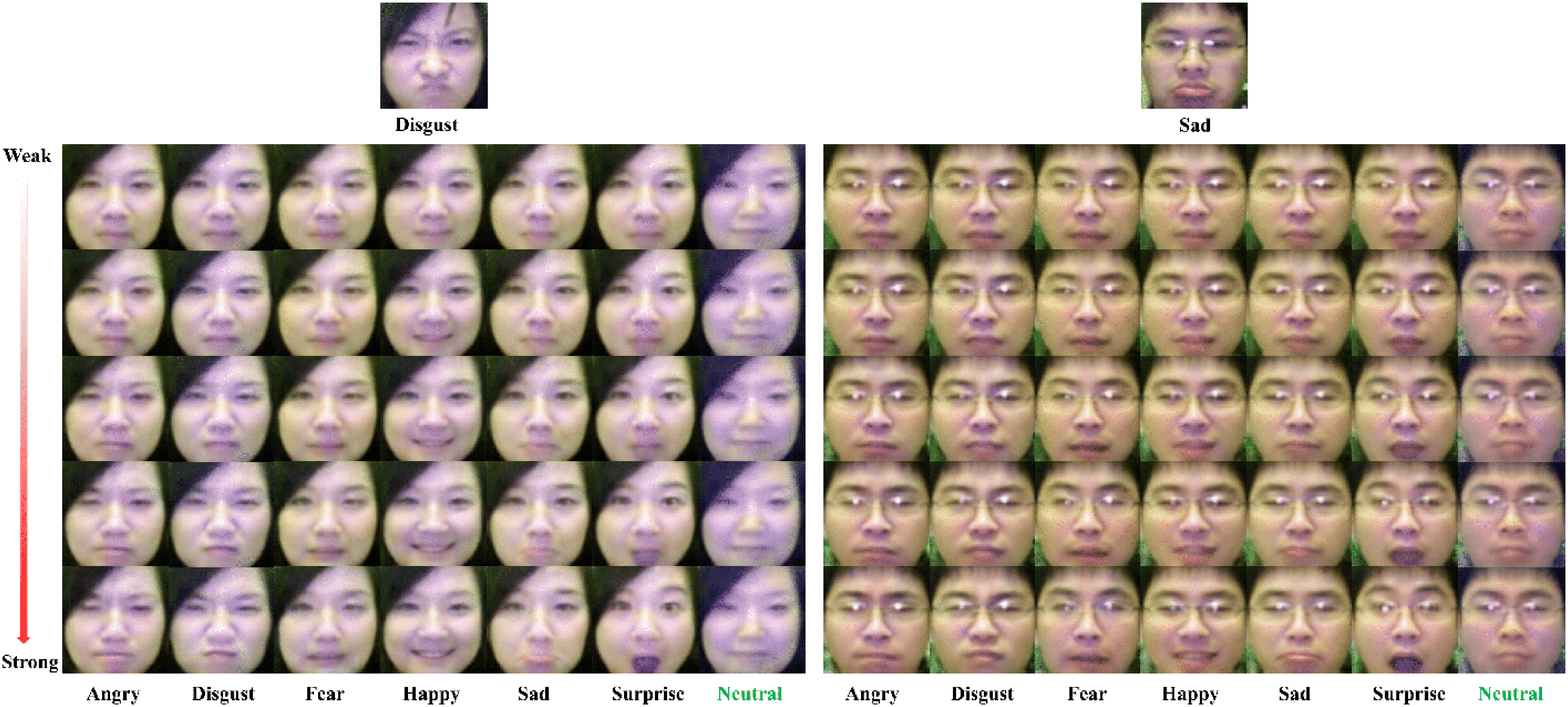}\label{subfig:exprgansample}}
    
    \subfigure[StarGAN]{\includegraphics[width=.41\textwidth]{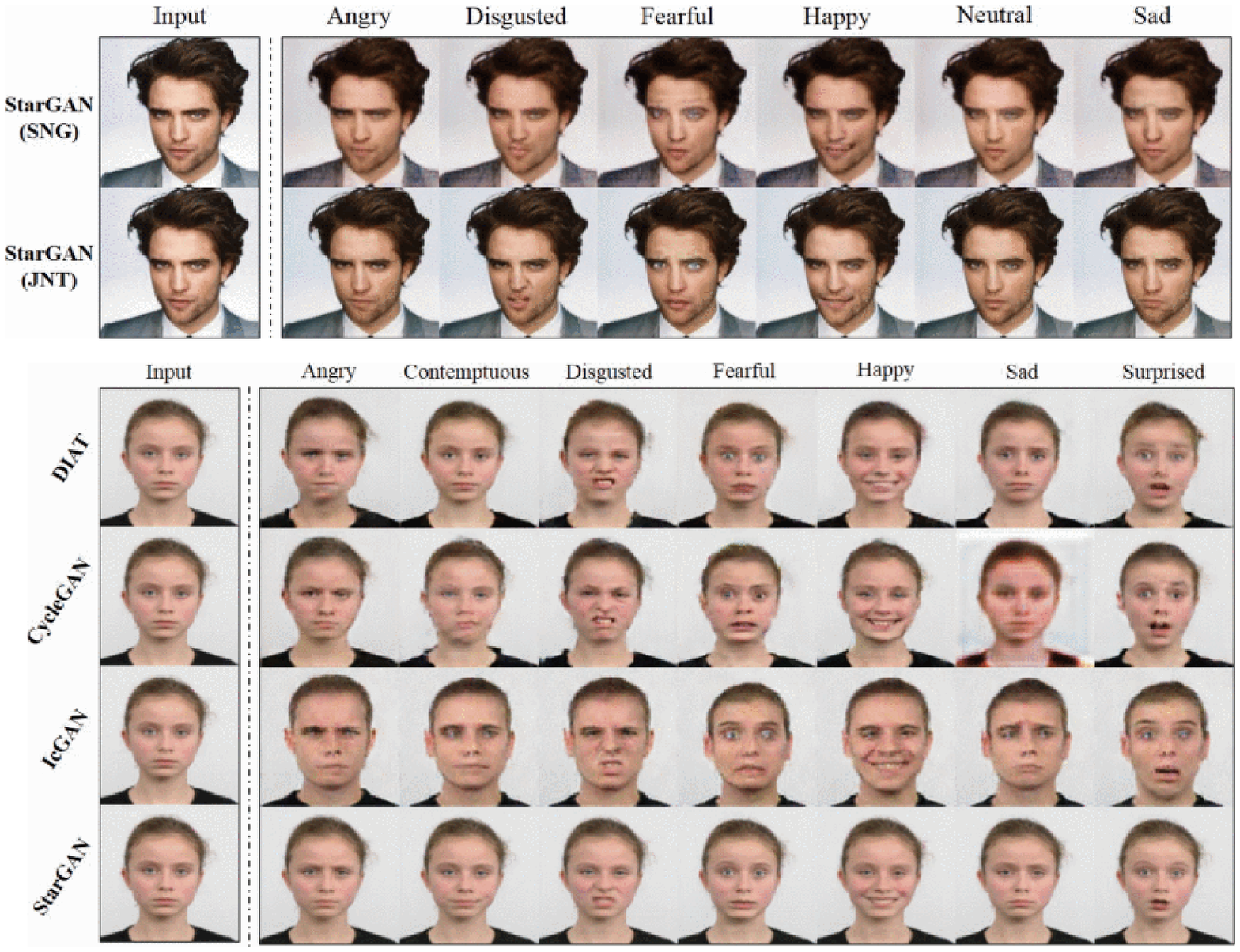}\label{subfig:stargansample}}\qquad   \subfigure[G2GAN]{\includegraphics[width=.4\textwidth]{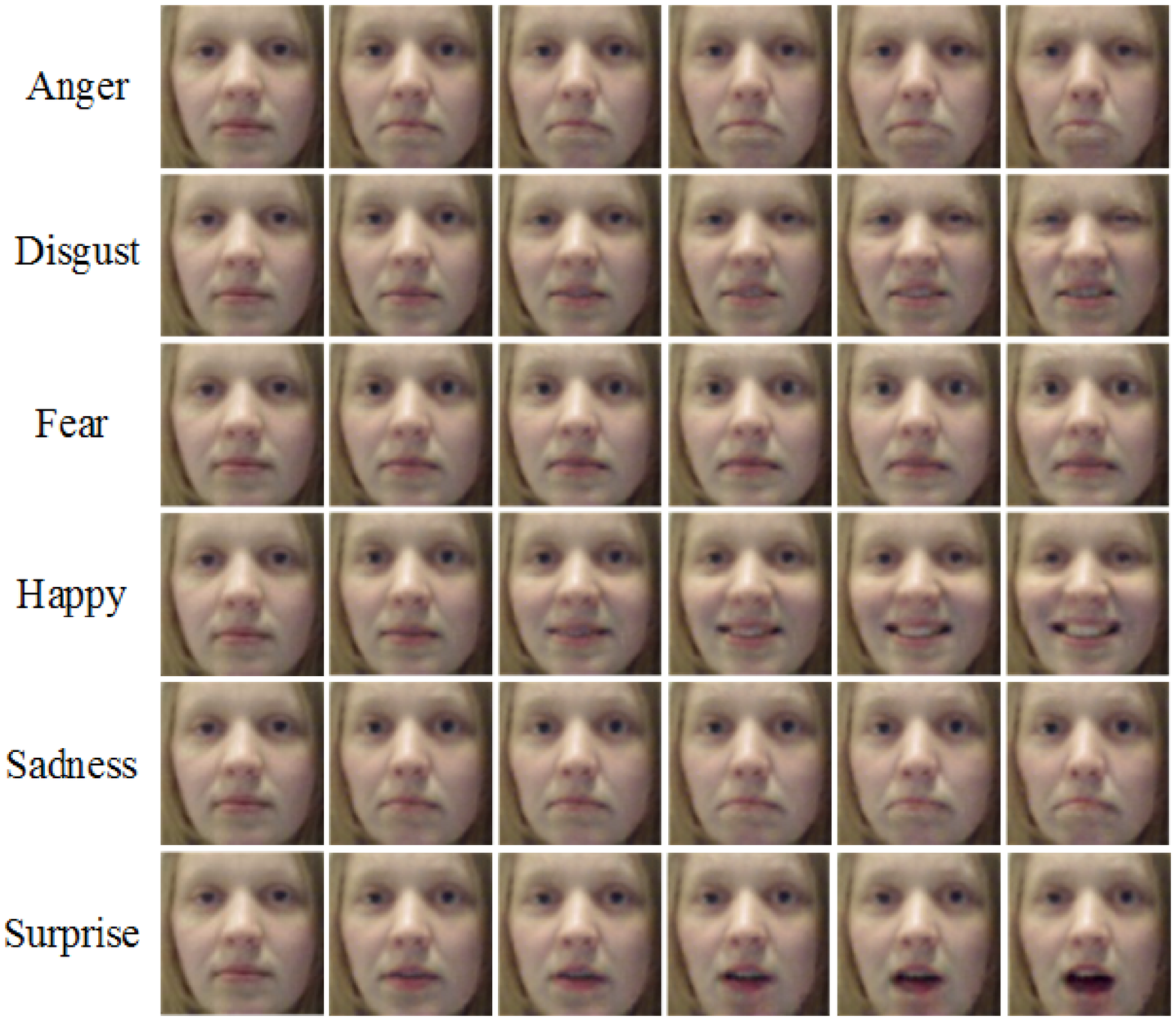}\label{subfig:g2gansample}}
    
    \subfigure[ \citeauthor{zhou2017photorealistic}]{\includegraphics[width=.45\textwidth]{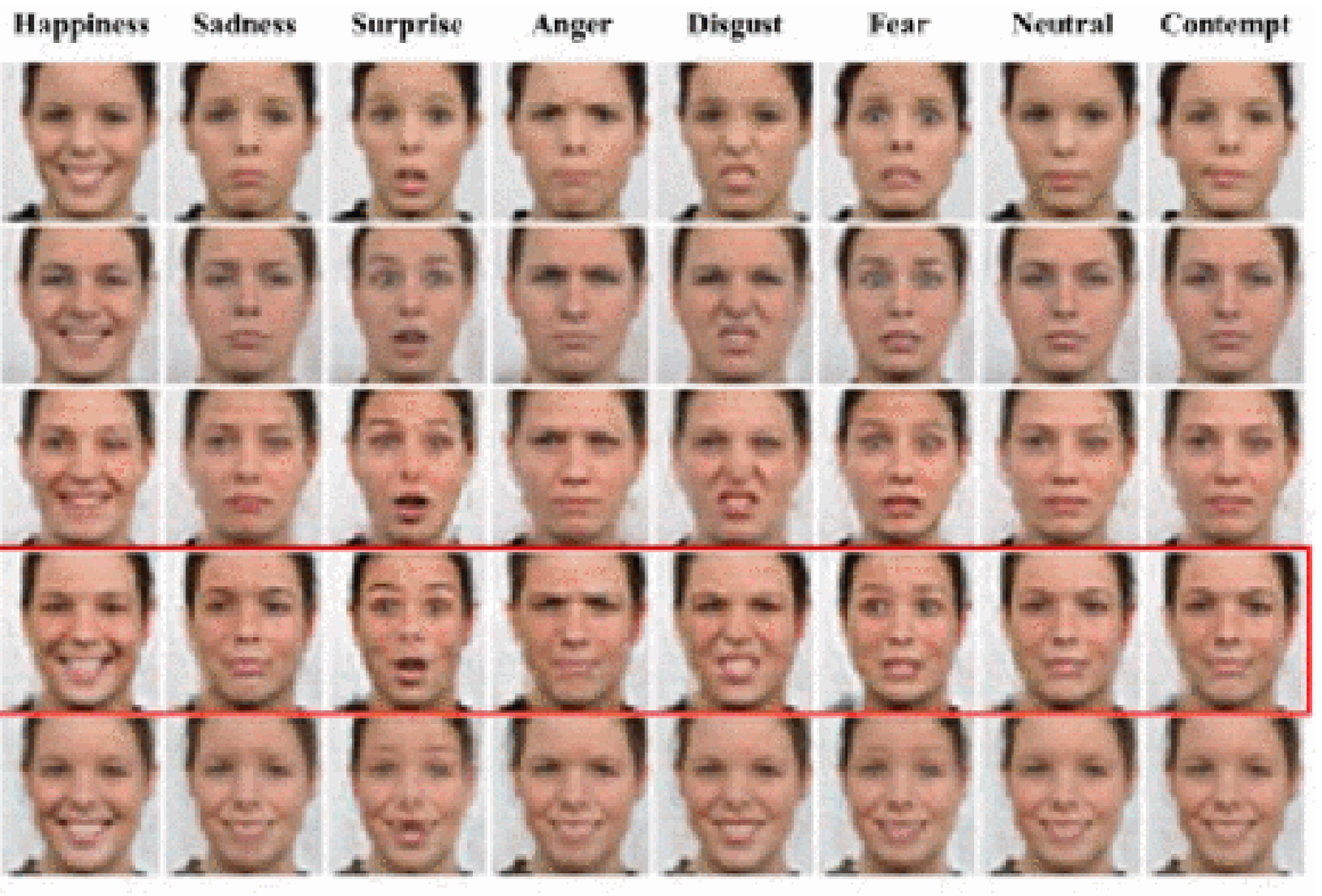}\label{subfig:zhugansample}}\qquad 
    \subfigure[FaceFeat]{\includegraphics[width=.31\textwidth]{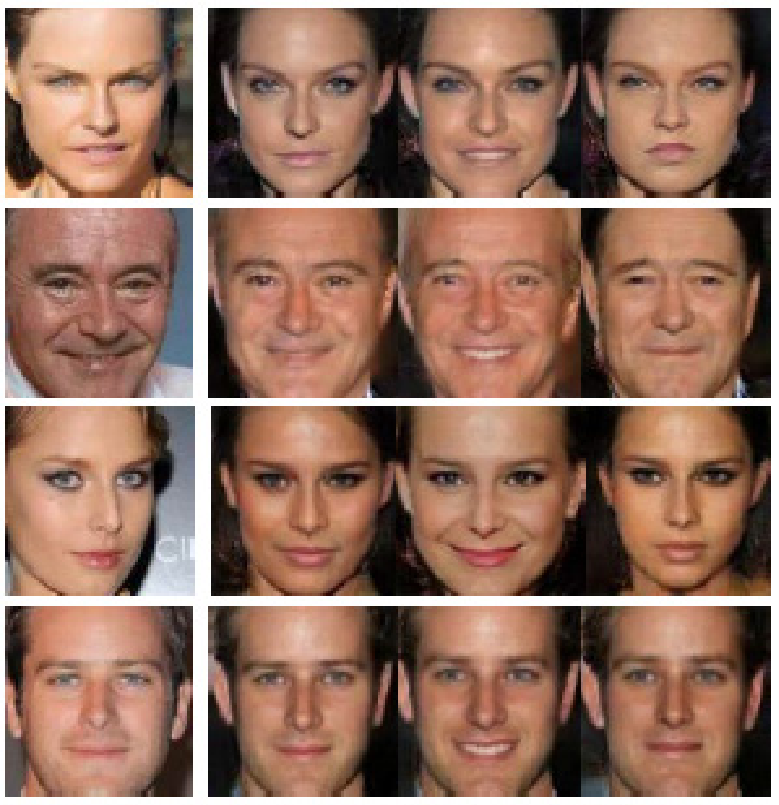}\label{subfig:facefeatgansample}}  
    
    \caption{Visual comparison of the GAN models, images are in courtesy of the reviewed papers }\label{fig:gansamples1}
\end{figure}

\begin{figure}[h] 
    \subfigure[DyadGAN, top to bottom: Joy, Anger, Surprise, Fear, Contempt, Disgust, Sad and Neutral]{\includegraphics[width=.7\textwidth]{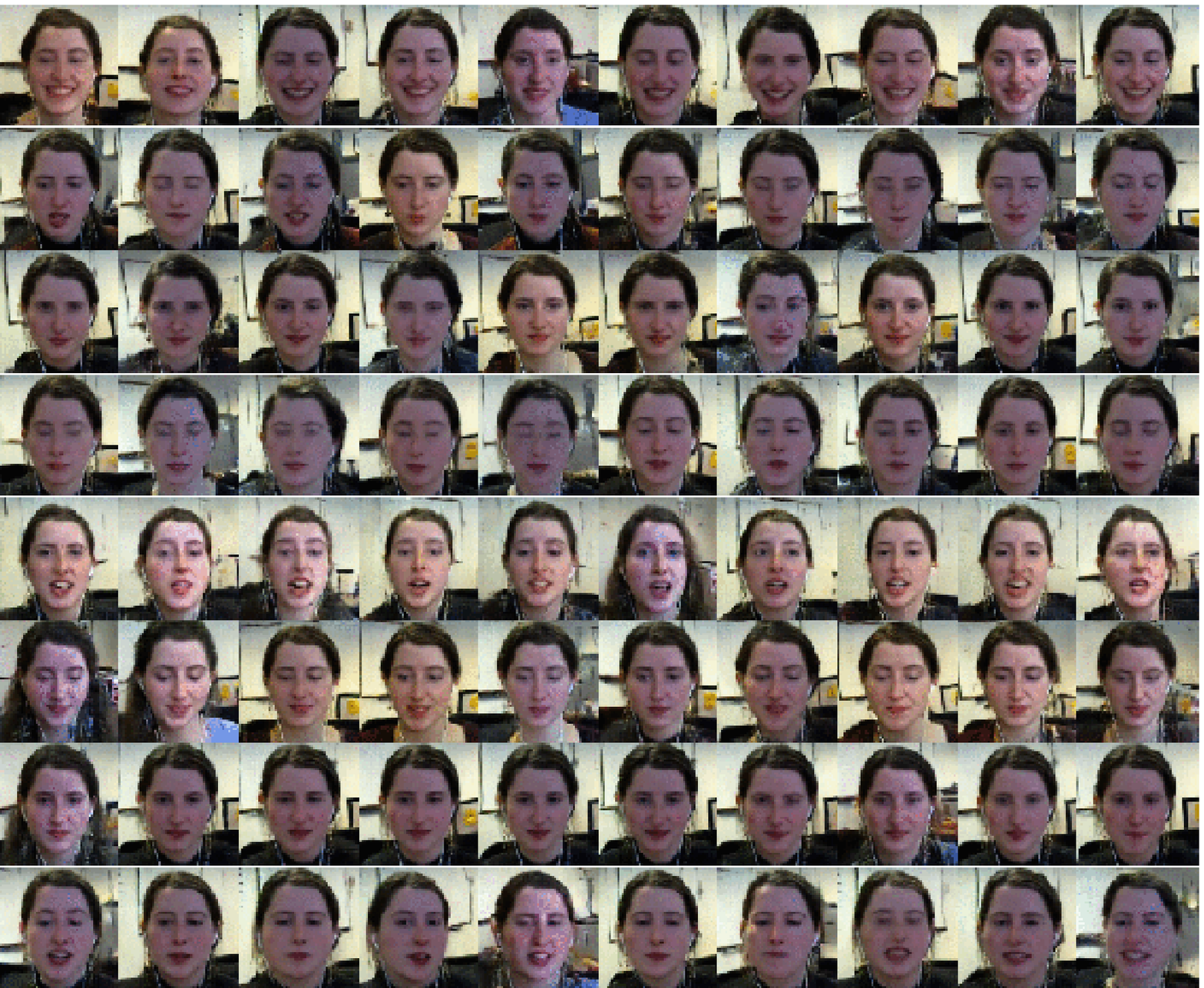}\label{subfig:dyadgansample}}
    
    \subfigure[ExprADA, left to right: input face, angry, disgusted, fearful, happiness, neutral, sadness and surprised , respectively.]{\includegraphics[width=.7\textwidth]{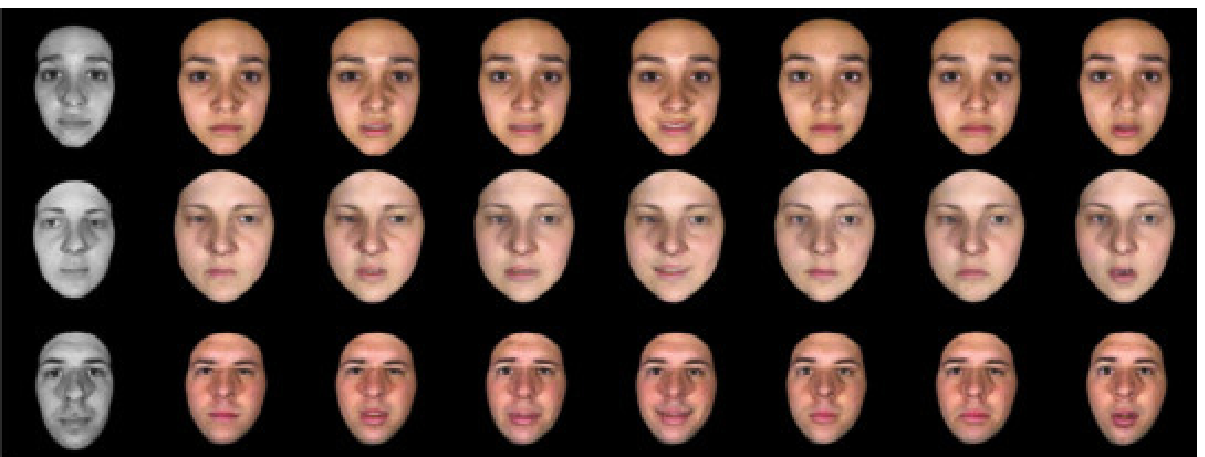}\label{subfig:expradasample}}
    
    \subfigure[CollaGAN]{\includegraphics[width=.7\textwidth]{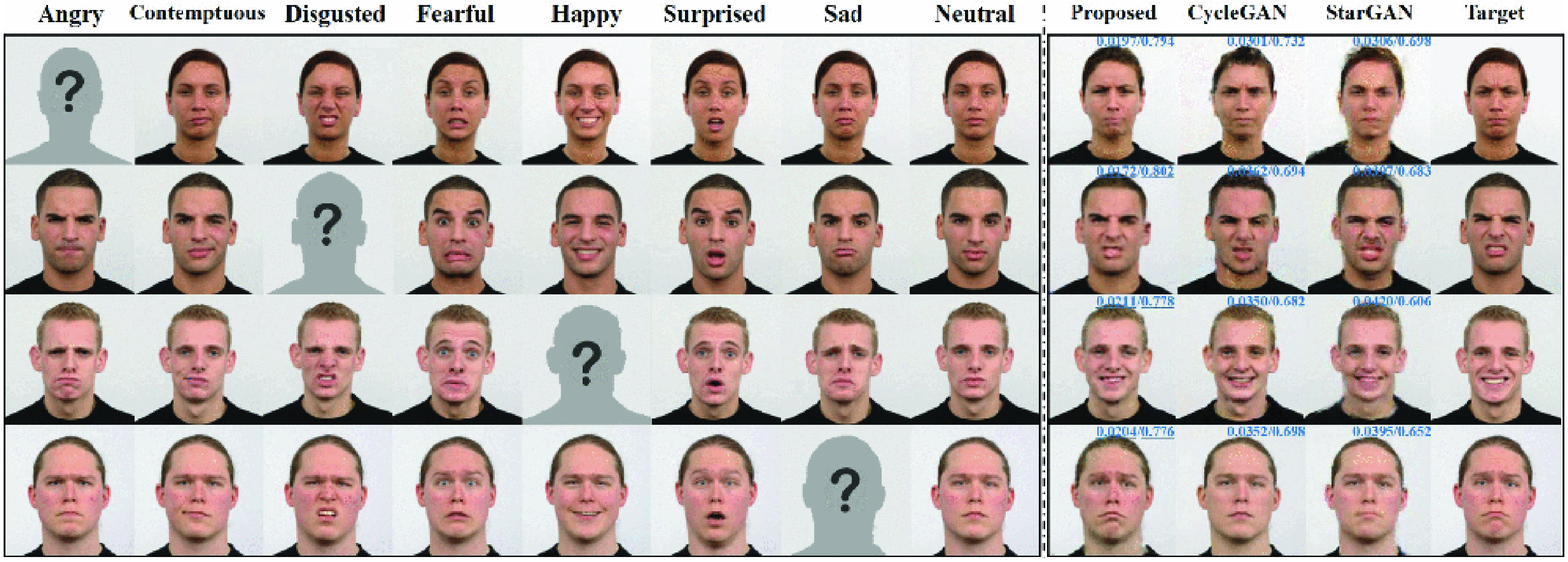}\label{subfig:collagansample}}
    
    \caption{Visual comparison of the GAN models, images are in courtesy of the reviewed papers }\label{fig:gansamples2}
\end{figure}

It should be noted that a series of GAN models focus on 3D object/face generation. Examples of these models are Convolutional Mesh Autoencoder (CoMA) \citep{ranjan2018generating}, MeshGAN\citep{cheng2019meshgan}, UVGAN \citep{deng2018uv}, and MeshVAE \citep{litany2018deformable}. Despite the successful performance of GANs in image synthesis, they still fall short when dealing with 3D objects and particularly human face synthesis. Here, we compare synthesized images of the aforementioned methods qualitatively in Figures \ref{fig:gansamples1} and \ref{fig:gansamples2}. Images are taken from the corresponding papers. As the images show, most of the generated samples suffer from blurring problem.
 
In addition to GAN-based models that synthesize single images, there are models with the ability to generate an image sequence or a video/animation. Video GAN (VGAN) \citep{vondrick2016generating} and Temporal GAN (TGAN) \citep{saito2017temporal} were the first two models in this research line. Although these models could learn a semantic representation of unlabeled videos, they produced a fixed-length video clip. As a result, MoCoGAN is proposed by \citeauthor{tulyakov2018mocogan} to solve the problem. MoCoGAN is composed of 4 sub-networks. These sub-networks are a recurrent neural network, an image generator, an image discriminator, and a video discriminator. The image generator generates a video clip by sequentially mapping a sequence of vectors to a sequence of images. 
{\renewcommand{\arraystretch}{1.4}
\begin{table*}[h]
    \setlength{\aboverulesep}{0pt}
    \setlength{\belowrulesep}{0pt}
    \fontsize{8}{6.5}\selectfont
    \caption{Comparison of facial expression video generation models, description of loss functions (L), metrics (M), databases (D) and purposes (P) used in the reviewed publications are given in Tables \ref{tab:fdatabases}, \ref{tab:flosses}, \ref{tab:fmetrics},  and \ref{tab:fpur}} \label{tab:FacialVideoGANS}
    \begin{tabular}{L{1.4cm}L{1.5cm}L{1.9cm}L{1cm}L{.6cm}L{.7cm}L{.85cm}L{.8cm}}
        \toprule
        author & based on & method & loss & data & M & RS & RM\\
        \toprule
        \makecell[l]{Pumarola \\ et al.} & CycleGAN & GANimation & \makecell[l]{L$v_{1}$,\\ L$v_{3}$,\\ L$v_{5}$,\\ L$v_{11}$,\\ L$v_{12}$} & D$v_{7}$ & M$v_{7}$ & - & \makecell[l]{P$v_{6}$,\\ P$v_{9}$,\\ P$v_{15}$}\\
        \midrule
        \multirow{6}{*}{\makecell[l]{Tulyakov \\ et al.}}& \multirow{6}{*}{CGAN} & MoCoGAN* & \multirow{6}{*}{\makecell[l]{L$v_{1}$,\\ L$v_{7}$}} & \multirow{6}{*}{D$v_{14}$} & \multirow{3}{*}{M$v_{12}$} & 0.201 & \multirow{6}{*}{\makecell[l]{P$v_{4}$,\\ P$v_{7}$,\\ P$v_{14}$}}\\
        & & VGAN & & & & 0.322 & \\
        & & TGAN & & & & 0.305 & \\
        \cmidrule{6-7}
        & & MoCoGAN\textsuperscript{*} & & & M$v_{13}$ & 0.581 & \\
        \cmidrule{6-7}
        &  & */VGAN &  &  & \multirow{2}{*}{M$v_{8}$} & 84.20 & \\
        & & */TGAN & & & & 54.70 & \\
        \midrule
        \multirow{4}{*}{\citeauthor{qiao2018emotional}} & \multirow{4}{*}{VAEGAN} & \multirow{4}{*}{-} & \multirow{4}{*}{\makecell[l]{L$v_{1}$,\\ L$v_{9}$,\\ L$v_{10}$}} & \multirow{2}{*}{D$v_{2}$} & M$v_{5}$ & 0.69 & \multirow{4}{*}{\makecell[l]{P$v_{4}$,\\ P$v_{8}$,\\ P$v_{9}$\\ P$v_{19}$}}\\
        & & & & & M$v_{6}$ & 26.73 & \\
        \cmidrule{5-7}
        & & & & \multirow{2}{*}{D$v_{3}$} & M$v_{5}$ & 0.77 &\\
        & & & & & M$v_{6}$ & 27.67 & \\
        \midrule
        \makecell[l]{\\ \citeauthor{geng2018warp}} & \makecell[l]{\\ CGAN} & \makecell[l]{wg-GAN/ \\ \citeauthor{elor2017bringingPortraits}} & \makecell[l]{\\ L$v_{1}$} & D$v_{14}$ & \makecell[l]{\\ M$v_{8}$} & \makecell[l]{\\ 62.00} & \makecell[l]{P$v_{8}$,\\ P$v_{9}$, \\ P$v_{14}$,\\ P$v_{19}$} \\
        \midrule
        \makecell[l]{Nakahira \\ and\\ Kawamoto}& CGAN & DCVGAN & L$v_{1}$ & D$v_{14}$ & M$v_{14}$ & 6.68 & P$v_{15}$ \\
        \midrule
        \multirow{5}{*}{\makecell[l]{Yang\\ et al.}} & \multirow{5}{*}{-} & PS/SCGAN* & \multirow{5}{*}{\makecell[l]{L$v_{1}$,\\ L$v_{10}$,\\ L$v_{12}$}} & \multirow{5}{*}{D$v_{3}$} & \multirow{3}{*}{M$v_{14}$} & 1.92 & \multirow{5}{*}{\makecell[l]{P$v_{1}$,\\ P$v_{15}$}} \\
         & & VGAN & & & & 1.68 & \\
         & & MoCoGAN & & & & 1.83 & \\
         \cmidrule{6-7}
         & & */VGAN & & &\multirow{2}{*}{M$v_{8}$} & 93.00 & \\
         & & */MoCoGAN & & & & 86.00 & \\
         \midrule
        \citeauthor{kim2018deep} & GAN & DVP/VDub & L$v_{1}$ & - & M$v_{8}$ & 51.25 & \makecell[l]{P$v_{4}$,\\ P$v_{5}$,\\ P$v_{15}$} \\
         \midrule
        \citeauthor{bansal2018recycle} & CycleGAN & RecycleGAN & L$v_{1}$ & - & M$v_{8}$ & 76.00 & \makecell[l]{P$v_{10}$,\\ P$v_{15}$,\\ P$v_{21}$,\\ P$v_{22}$} \\
        \midrule
       \multirow{2}{*}{\citeauthor{sun2020twostreamvan}}& \multirow{2}{*}{CGAN} & 2SVAN\textsuperscript{*$\dagger$} & \makecell[l]{L$v_{1}$,\\ L$v_{7}$,} & \multirow{2}{*}{D$v_{14}$} & M$v_{14}$ & 5.48 & \makecell[l]{P$v_{8}$,\\ P$v_{9}$,} \\
        & & */MoCoGAN & \makecell[l]{L$v_{21}$,\\ L$v_{22}$} & & M$v_{8}$ & 88.00 & \makecell[l]{P$v_{14}$,\\ P$v_{19}$}\\
        \bottomrule
    \end{tabular}
    \begin{flushleft}
        \footnotesize{
        - M: Metric, RS: Results, RM:Remarks \\
        - *: shows the proposed method by authors, other mentioned methods are implemented by the authors for the sake of comparison \\
        - the result reported for expression classification accuracy (M$v_{2}$) belongs to the synthesized image datasets \\
        - All papers provide visual representation of the synthesized images (M$v_{7}$)\\
        - \textsuperscript{*$\dagger$}:TwoStreamVAN
        }
    \end{flushleft}
 \end{table*}
}

While MoCoGAN uses content and motion, Depth Conditional Video generation (DCVGAN) proposed by \citet{nakahira2019dcvgan} utilizes both the optical information and the 3D geometric information to generate accurate videos using the scene dynamics. DCVGAN solved the unnatural appearance of moving objects and assimilation of objects into the background in MoCoGAN. Other methods like Warp-guided GAN \citep{geng2018warp} generate real-time facial animations using a single photo. The method instantly fuses facial details like wrinkles and creases to achieve a high fidelity facial expression.

Recently, \citeauthor{yang2018pose} (\citeyear{yang2018pose}) proposed a pose-guided method to synthesize human videos. This successful method relies on two concepts: first, a Pose Sequence Generative Adversarial Network (PSGAN) is proposed to learn various motion patterns by conditioning on action labels. Second, a Semantic Consistent Generative Adversarial Network (SCGAN) is employed to generate image sequences (video) given the pose sequence generated by the PSGAN. The effect of noisy or abnormal poses between the generated and ground-truth poses is reduced by the semantic consistency. We show this method as PS/SCGAN in Table \ref{tab:FacialVideoGANS}. It is worth to mention that two of the recent and successful methods in video generation are MetaPix \citep{lee2019metapix} and MoCycleGAN \citep{chen2019mocycle} that used motion and temporal information for realistic video synthesis. However, these methods are not tested for facial expression generation. Table \ref{tab:FacialVideoGANS} lists the models developed for video or animation generation.

One of the main goals in synthesizing is augmenting the number of available samples. \citet{zhu2018emotion} used GAN models to improve the imbalanced class distribution by data augmentation through GAN models. The discriminator of the model is a CNN and the generator is based on CycleGAN. They report up to 10\% increase in the classification accuracy (M$v_{2}$) based on GAN-based data augmentation techniques.

\begin{figure}[b] 
    \centering
    \subfigure[Ganimation]{\includegraphics[width=.8\textwidth]{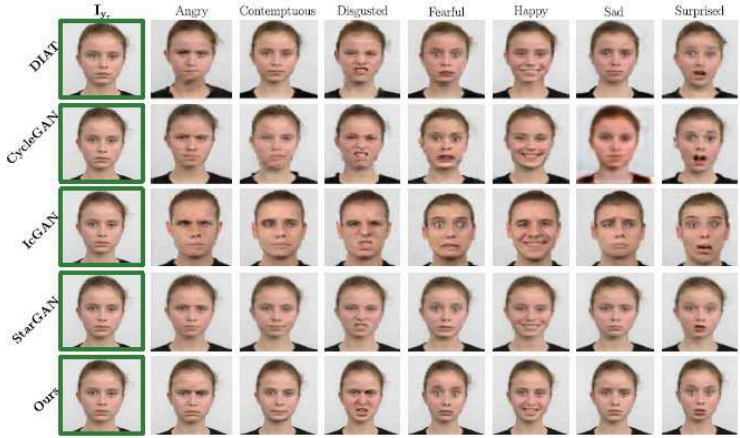}\label{subfig:ganimationsample}}
    
    \subfigure[MocoGAN]{\includegraphics[width=.89\textwidth]{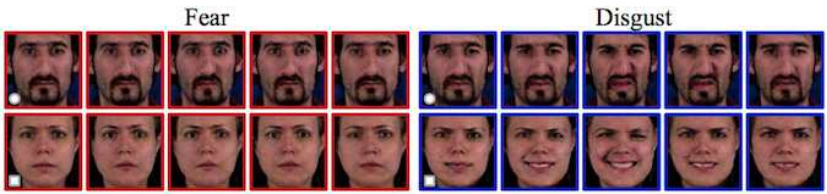}\label{subfig:mocogansample}}
    
    \caption{Visual comparison of the GAN models, images are in courtesy of the reviewed papers }\label{fig:gansamples3}
\end{figure}

The objective function or the optimization loss problem categorizes into two groups: synthesis loss and classification loss. Although the definitions provided by the authors are not always clear, we tried to list all different losses used by authors and we propose a symbolic name for each to provide harmony in the literature. The losses are used in a general point of view. That is, marking different papers by classification loss (L7) in Table \ref{tab:FacialImageGANS}, does not mean necessarily that the exact same loss function is used. In other words, it shows that the classification loss is contributed in some way. A comprehensive list of these functions is given in Table \ref{tab:flosses}. Additionally, we compared some of the video synthesis models in Figure \ref{fig:gansamples3}.

Evaluation metrics of the generative models are different from one research to another due to several reasons \citep{hitawala2018comparative}. First, the quality of the synthesized sample is a perceptual concept and, as a result, it cannot be accurately expressed. Usually, researchers provide the best-synthesized samples for visual comparison and thus problems like mode drop are not covered qualitatively. Second, employing human annotators to judge the visual quality can cover only a limited number of data samples. Specifically, in topics such as human emotion,  experts are required for accurate annotation and having the least possible labeling error. Hence, approaches like Amazon Mechanical Turk are less reliable considering classification based on those labels. Third, general metrics like photo-metric error, geometric error, and inception score are not reported in all publications \citep{salimans2016improved}. These problems cause the comparison among papers either unfair or impossible. 


The Inception Score (IS) can be computed as follows:
\begin{equation}
    \text{IS} = \exp(\mathbb{E}_{\x_g}[\mbox{KL}(p(y|x_g)\mid\mid p(y))]),
\end{equation}
where $x_g$ denotes the generated sample, y is the label predicted by an arbitrary classifier, and $\mbox{KL}(.)$ is the KL divergence to measure the distance between probability distributions as defined in Eq. (\ref{kld}). Based on this score, an ideal model produces samples that have close congruence to real data samples as much as possible. In fact, KL divergence is the de-facto standard for training and evaluating generative models. 

Other widely used evaluative metrics are Structural Similarity Index Measure (SSIM) and Peak Signal to Noise Ratio (PSNR). SSIM is expressed as follows:

\begin{equation}
    \text{SSIM}(x,y) = I(x,y)^\alpha C(x,y)^\beta S(x,y)^\gamma ,
\end{equation}
where $I$, $C$, and $S$ are luminance, contrast, and structure and they can be formulated as:

\begin{equation}
    I(x,y)= \frac{2\mu_x\mu_y+C_1}{\mu_x^2+\mu_y^2+C_1} \quad C(x,y)=\frac{2\sigma_x\sigma_y+C_2}{\sigma_x^2+\sigma_y^2+C_2} \quad S(x,y)=\frac{\sigma_{xy}+C_3}{\sigma_x\sigma_y+C_3}
\end{equation}

Here $\mu_x$, $\mu_y$, $\sigma_x$, and $\sigma_y$ denote mean and standard deviations of pixel intensity in a local image patch where the patch is superimposed so that its center coincides with the center of the image. Typically, a patch is considered as a square neighborhood of $n \times x$ pixels. Also, $\sigma_{xy}$ is the sample correlation coefficient between corresponding pixels in that patch. $C1$, $C2$, and $C3$ are small constants values added for numerical stability.

PSNR or the peak signal-to-noise ratio assesses the quality between two monochrome images $\x_g$ and $\x_r$. Let $\x_g$ and $\x_r$ be the generated image and the real image, respectively. Then, PSNR is:

\begin{equation}
    \text{PSNR}(\x_g,\x_r)=20 \log_{10}\bigg(\frac{\text{MAX}_{\x_r}}{\text{MSE}(\x_g,\x_r)}\bigg),
\end{equation}
where MAX$_{\x_r}$ is the maximum possible pixel value of the image and MSE stands for Mean Square Error. PSNR is measured in dB, generated images with a better quality result in higher PSNR.

In addition to the metrics that evaluate the generated image, Generative Adversarial Metric (GAM) proposed by \citeauthor{im2016generating} (\citeyear{im2016generating}) compares two GAN models by engaging them in a rivalry. In this metric, first GAN models $M_1$ and $M_2$ are trained. Then, model $M_1$ competes with model $M_2$ in a test phase by having $M_1$ trying to fool discriminator of $M_2$ and vice versa. In the end, two ratios are calculated using the discriminative scores of these models as follows:
\begin{equation}
    r_{\text{test}} \stackrel{\text{def}}{=} \frac{\epsilon(D_1(X_{\text{test}}))}{\epsilon(D_2(X_{\text{test}}))} \text{\quad, and\quad} 
    r_{\text{sample}} \stackrel{\text{def}}{=} \frac{\epsilon(D_1(G_2(Z)))}{\epsilon(D_2(G_1(Z)))}
    \label{eq:GAMratio}
\end{equation}
where $G_1$, $D_1$, $G_2$, and $D_2$, are the generators and the discriminators of $M_1$ and $M_2$, respectively. In Eq. (\ref{eq:GAMratio}), $\epsilon(.)$ outputs the classification error rate. The test ratio or $r_{\text{test}}$ shows which model generalizes better because it discriminates based on $X_{\text{test}}$. The sample ratio or $r_{\text{sample}}$ shows which model fools the other more easily because discriminators classify based on the synthesized samples of the opponent. The sample ratio and the test ratio can be used to decide the winning model:
\begin{equation}
    \text{winner} = 
    \begin{cases}
    M_1 \text{\quad if $r_{\text{sample}} < 1$ and $r_{\text{\text{test}}}  \simeq 1$ } \\
    M_2 \text{\quad if $r_{\text{sample}} > 1$ and $r_{\text{\text{test}}} \simeq 1$ }  \\
    \text{Tie} \text{\quad otherwise}
    \end{cases}
\end{equation}

To measure the texture similarity, \citet{peng2019apprgan} simply calculated correlation coefficients between $T_g$ and $T_r$ that are the texture of the synthesized image and the texture of its corresponding ground truth, respectively. Let $\rho$ be the texture similarity score. Then, the mathematical representation is as follows:
\begin{equation}
    \rho = \frac{\sum_i \sum_j(T_r(i,j)-\mu_r)(T_g(i,j)-\mu_g)}{\sqrt{\sum_i \sum_j(T_r(i,j)-\mu_r)^2\sum_i\sum_j(T_g(i,j)-\mu_g)^2}},
\end{equation}
where (i, j) specifies pixel coordinates in the texture images, and $\mu_g$ and $\mu_r$ are the mean value of $T_g$ and $T_r$, respectively. 

Other important metrics include Fréchet Inception Distance (FID), Maximum Mean Discrepancy (MMD), the Wasserstein Critic, Tournament Win Rate and Skill Rating, and Geometry Score. FID works based on embedding the set of synthesized samples into a new feature space using a certain layer of a CNN architecture. Then, mean and covariance are estimated for both the synthesized and the real data distributions based on the assumption that the embedding layer is a continuous multivariate Gaussian distribution. Finally, FID or Wasserstein-2 distance between these Gaussians is then used to quantify the quality of generated samples :
\begin{equation}
    \text{FID}(r,g) = \parallel \mu_r - \mu_g \parallel_2^2 + \text{Tr}(\sum_r + \sum_g - 2(\sum_r\sum_g)^{\frac{1}{2}}).
\end{equation}

Here, $(\mu_g, \sum_g)$ and $(\mu_r, \sum_r)$ represent the mean and covariance of generated and real data distributions, respectively. Lower FID score indicates a smaller distance between the two distributions. MMD focuses on the dissimilarity between the two probability distributions by taking samples from each distribution independently. The kernel MMD is expressed as follows:
\begin{align}
    M_K(P_r,P_g) =& \mathbb{E}_{\x_r,\x_r' \sim P_r}[k(\x_r,\x_r')] \nonumber \\ + & \mathbb{E}_{\x_g, \x_g' \sim P_g}[k(\x_g, \x_g')] \nonumber \\ - & 2\mathbb{E}_{\x_r\sim P_r, \x_g\sim P_g}[k(\x_r,\x_g)]
\end{align}
where $k$ is some fixed characteristic kernel function like Gaussian kernel: $k(x_r, x_g) = \text{exp}(\parallel x_r - x_g \parallel^2)$ that measures MMD dissimilarity between the generated and real data distributions. Also, $\x_r$ and $\x_r'$ are randomly drawn samples from real data distribution, i.e $P_r$. Similarly, $\x_g$ and $\x_g'$ are randomly drawn from model distribution, i.e $P_g$.

The Wasserstein Critic provides an approximation
of the Wasserstein distance between the model distribution and the real data distribution. Let $P_r$ and $P_g$ be the real data and the model distributions, then:
\begin{equation}
    W(P_r, P_g) \propto \max_f \mathbb{E}_{\x_r \sim P_r}[f(\x_r)] - \mathbb{E}_{\x_g \sim P_g}[f(\x_g)],
\end{equation}

where $f: \mathbb{R}^D \longrightarrow \mathbb{R} $ is a Lipschitz continuous function. In practice, the critic $f$ is a neural network with clipped weights and bounded derivatives \citep{borji2019pros}. In practice, this is approximated by training to achieve high values for real samples and low values for generated ones:

\begin{equation}
    \hat{W}(X_{\text{test}},X_g) = \frac{1}{N}\sum_{i=1}^{N}\hat{f}(X_{\text{test}}[i]) - \frac{1}{N}\sum_{i=1}^{N}\hat{f}(X_g[i]),
\end{equation}
where $X_{\text{test}}$ is a batch of testing samples, $X_g$ is a batch of generated samples, and $\hat{f}$ is the independent critic. An alternative version of this score is known as Sliced Wasserstein Distance (SWD) that estimates the Wasserstein-1 distance (see Eq. (\ref{wass})) between real and generated images. SWD computes the statistical similarity between local image patches extracted from Laplacian pyramid representations of the images \citep{karras2017progressive}.

In the case of the metrics of video generation, evaluating content consistency based on Average Content Distance (ACD) is defined as calculating the average pairwise $L_2$ distance of the per-frame average feature vectors. In addition, the motion control score (MCS) is suggested for assessing the motion generation ability of the model. Here, a spatio-temporal CNN is first trained on a training dataset. Then, this model classifies the generated videos to verify whether the generated video contained the required motion (e.g action/expression). 

Other metrics include but are not limited to identification classification, true/false acceptance rate \citep{song2018geometry}, expression classification accuracy/error \citep{ding2018exprgan}, real/fake classification accuracy/error \citep{ding2018exprgan}, attribute editing accuracy/error \citep{he2019attgan}, and Fully Convolutional Networks. List of evaluative metrics used in the reviewed publications is given in Table \ref{tab:fmetrics}. For a comprehensive list on evaluative metrics of GAN models, we invite the reader to study “Pros and Cons of GAN Evaluation Measures" by \cite{borji2019pros}.

Synthesizing models are proposed with different aims and purposes. Texture synthesis, super-resolution images, and image in-painting are some applications. Considering face synthesis, the most important goal is the data augmentation for improved recognition performance. A complete list of such purposes and the model properties are given in Table \ref{tab:fpur}. 
\begin{table}[t]
    \setlength{\aboverulesep}{0pt}
    \setlength{\belowrulesep}{0pt}
    \caption{List of purposes and characteristics of GAN models used for facial emotion synthesis by the reviewed publications }
    \label{tab:fpur}
    \begin{tabular}{lL{10.55cm}}
        \toprule
        & purpose or characteristic\\
        \toprule
        P$v_1$ & is tested for data augmentation \\
        P$v_2$ & preserves the identity of the subject\\
        P$v_3$ & controls the expression intensity \\
        P$v_4$ & generates images with high quality (photo-realistic)\\
        P$v_5$ & employs geometrical features as the conditional vector $c$\\
        P$v_6$ & employs facial action units as the conditional vector $c$\\
        P$v_7$ & employs a unified framework for multi-domain tasks \\
        P$v_8$ & generates facial expressions by combining appearance and geometric features \\
        P$v_9$ & provides a smooth transition of facial expression \\
        P$v_{10}$ & supports training with unpaired samples \\
        P$v_{11}$ & employs arousal-valence and dominance-like inputs\\
        P$v_{12}$ & preserves the emotion of the subject\\
        P$v_{13}$ & employs arbitrary head poses and applies face frontalization\\
        P$v_{14}$ & modifies facial attributes based on a desired value\\
        P$v_{15}$ & generates image sequences (video)\\
        P$v_{16}$ & employs multiple inputs\\
        P$v_{17}$ & modifies image attributes like illumination and background \\
        P$v_{18}$ & is tested for data imputation \\
        P$v_{19}$ & generates video animation using a single image\\
        P$v_{20}$ & generates visible faces from thermal face image\\
        P$v_{21}$ & employs temporal/motion information for video generation \\
        P$v_{22}$ & supports unsupervised learning \\
        \bottomrule
    \end{tabular}
\end{table}

Despite the numerous publications on image and video synthesis, yet some problems are not solved thoroughly. For example, generating high-resolution samples is an open research problem. The output is usually blurry or impaired by checkered artifacts. Results obtained for video generation or synthesis of 3D samples are far from realistic examples. Also, it is important to highlight that the number of publications focused on expression classification is greater than that of those employing identity recognition. 

\subsection{Speech Emotion Synthesis}
Research efforts focusing on synthesizing speech with emotion effect has continued for more than a decade now. One application of GAN models in speech synthesis is speech enhancement. A pioneer GAN-based model developed for raw speech generation and enhancement is called the Speech Enhancement GAN (SEGAN) proposed by \cite{pascual2017segan}. SEGAN provides a quick non-recursive framework that works End-to-End (E2E) with raw audio. Learning from different speakers and noise types and incorporating that information to a shared parameterizing system is another contribution of the proposed model. Similar to SEAGAN, \cite{macartney2018improved} proposes a model for speech enhancement based on a CNN architecture called Wave-UNet. The Wave-UNet is used successfully for audio source separation in music and speech de-reverberation. Similar to section \ref{subsec:fes}, we compare the results of the reviewed papers in Table \ref{tab:RawSpeechGANS}. Additionally, Tables \ref{tab:sdatabases} to \ref{tab:spur} represent databases, loss functions, assessment metrics and characteristics used in speech synthesis. 

\cite{sahu2018enhancing} followed a two-fold contribution. First, they train a simple GAN model to learn a high-dimensional feature vector through the distribution of a lower-dimensional representation. Second, cGAN is used to learn the distribution of the high-dimensional feature vectors by conditioning on the emotional label of the target class. Eventually, the generated feature vectors are used to assess the improvement of emotion recognition. They report that using synthesized samples generated by cGAN in the training set is helpful. Also it is concluded that using synthesized samples in the test set suggests the estimation of a lower-dimensional distribution is easier than a high-dimensional complex distribution. Employing the synthesized feature vectors from IEMOCAP database in a cross-corpus experiment on emotion classification of MSP-IMPROV database is reported to be successful.

Mic2Mic \citep{mathur2019mic2mic} is another example of a GAN-based model for speech enhancement. This model addresses a challenging problem called microphone variability. The Mic2Mic model disentangles the variability problem from the downstream speech recognition task and it minimizes the need for training data. Another advantage is that it works with unlabeled and unpaired samples from various microphones. This model defines microphone variability as a data translation from one microphone to another for reducing domain shift between the train and the test data. This model is developed based on CycleGAN to assure that the audio sample \citep{mathur2019mic2mic} from microphone A is translated to a corresponding sample from microphone B. 

\begin{center}
    \fontsize{8}{8}\selectfont
    \begin{longtable}{L{1.8cm}L{1.5cm}L{1.75cm}L{.7cm}L{.7cm}L{.8cm}L{.85cm}L{.8cm}}
    \toprule
        author & based on & model & loss & data & M & RS & RM\\
    \toprule \\
    \endfirsthead
    \multicolumn{8}{l}%
    {\tablename\ \thetable\ -- \textit{Continued from previous page}} \\
    \toprule
        author & based on & method & loss & data & M & RS & RM\\
    \toprule \\
    \endhead
    \hline \multicolumn{8}{c}{\textit{Continued on next page}} \\
    \endfoot
    \caption{Comparison of speech Emotion synthesis models, description of loss functions (L), metrics (M), databases (D) and purposes (P) used in the reviewed publications are given in Tables \ref{tab:sdatabases}, \ref{tab:slosses}, \ref{tab:smetrics},  and \ref{tab:spur}} \label{tab:RawSpeechGANS}\\
    \endlastfoot
        \multirow{6}{*}{\citeauthor{pascual2017segan}} & \multirow{6}{*}{CGAN} & \multirow{6}{*}{SEAGAN} & \multirow{6}{*}{L$a_{1}$} & \multirow{6}{*}{D$a_{1}$} & M$a_{1}$ & 2.16 & \multirow{6}{*}{P$a_{2}$} \\
        & & & & & M$a_{2}$ & 3.18 & \\
        & & & & & M$a_{3}$ & 3.48 & \\
        & & & & & M$a_{4}$ & 2.94 & \\
        & & & & & M$a_{5}$ & 2.80 & \\
        & & & & & M$a_{6}$ & 7.73 & \\
        \midrule
        \multirow{5}{*}{\makecell[l]{Macartney\\ and Weyde}} & \multirow{5}{*}{WaveUNet} & \multirow{5}{*}{-} & \multirow{5}{*}{L$a_{1}$} & \multirow{5}{*}{D$a_{1}$} & M$a_{2}$ & 2.41 & \multirow{5}{*}{P$a_{2}$} \\
        & & & & & M$a_{3}$ & 3.54 & \\
        & & & & & M$a_{4}$ & 3.24 & \\
        & & & & & M$a_{5}$ & 2.97 & \\
        & & & & & M$a_{6}$ & 9.98 & \\
        \midrule
        \citeauthor{mathur2019mic2mic} & CycleGAN & Mic2Mic & \makecell[l]{L$a_{1}$,\\ L$a_{2}$,\\ L$a_{6}$} & D$a_{8}$ & M$a_{8}$ & 89.00 & P$a_{2}$\\
        \midrule
        \citeauthor{hsu2017voice}& \makecell[l]{WGAN\\ CVAE\\ GAN} & VAWGAN & L$a_{1}$ & D$a_{2}$ & M$a_{2}$ & 3.00 & P$a_{3}$ \\
        \midrule
        \multirow{2}{*}{\citeauthor{latif2018adversarial}} & \multirow{2}{*}{GAN} &  \multirow{2}{*}{-} & \multirow{2}{*}{L$a_{1}$} & D$a_{3}$ &  \multirow{2}{*}{M$a_{7}$} & 47.87 & \multirow{2}{*}{\makecell[l]{P$a_{4}$,\\ P$a_{5}$}} \\
        &  &  &  & D$a_{4}$ & & 36.49 & \\
        \midrule
        \multirow{3}{*}{\makecell[l]{Kameoka \\ et al.}} &  \multirow{3}{*}{CVAE} & \multirow{3}{*}{CVAE-VC} & \multirow{3}{*}{\makecell[l]{L$a_{1}$,\\ L$a_{3}$,\\ L$a_{4}$}} & \multirow{3}{*}{D$a_{5}$} & M$a_{13}$ & 92.00 & \multirow{3}{*}{P$a_{5}$}\\
        & & & & & & & \\
        & & & & & M$a_{14}$ & 77.00 & \\
        \midrule
        \multirow{4}{*}{\makecell[l]{Kameoka\\ et al.}} &  \multirow{4}{*}{\makecell[l]{StarGAN\\ CycleGAN}} & \multirow{4}{*}{\makecell[l]{StarGAN-\\VC}} & \multirow{4}{*}{\makecell[l]{L$a_{1}$,\\ L$a_{2}$,\\ L$a_{3}$,\\ L$a_{6}$}} & \multirow{4}{*}{D$a_{5}$} & & & \multirow{4}{*}{P$a_{11}$}\\
        & & & & & M$a_{13}$ & 82.00 & \\
        & & & & & M$a_{14}$ & 67.00 & \\
        & & & & & & & \\
        \midrule
        \multirow{3}{*}{\makecell[l]{Kaneko and\\ Kameoka}} & \multirow{3}{*}{CycleGAN} & \multirow{3}{*}{\makecell[l]{CycleGAN-\\ VC}} & \multirow{3}{*}{\makecell[l]{L$a_{1}$,\\ L$a_{2}$,\\ L$a_{6}$}} & \multirow{3}{*}{D$a_{2}$} & M$a_{2}$ & 2.4 & \multirow{3}{*}{P$a_{3}$}\\
        & & & & & & & \\
        & & & & & M$a_{14}$ & 39.5 & \\
        \midrule
        \multirow{5}{*}{\citeauthor{kaneko2019cyclegan}} & \multirow{5}{*}{CycleGAN} & \multirow{5}{*}{\makecell[l]{CycleGAN-\\ VC2}} & \multirow{5}{*}{\makecell[l]{L$a_{1}$,\\ L$a_{2}$,\\ L$a_{6}$}} & \multirow{5}{*}{D$a_{5}$} & M$a_{2}$ & 3.1 & \multirow{5}{*}{\makecell[l]{P$a_{7}$\\ P$a_{10}$}}\\
        & & & & & M$a_{11}$ & 4.8 & \\
        & & & & & M$a_{13}$ & 69.5 & \\
        & & & & & M$a_{15}$ & 6.26 & \\
        & & & & & M$a_{16}$ & 1.45 & \\
        \midrule
        \citeauthor{tanaka2018synthetic}& \makecell[l]{CycleGAN\\ SEAGAN} & \makecell[l]{Wave-\\ CycleGAN-\\ VC} & \makecell[l]{L$a_{1}$,\\ L$a_{2}$} & D$a_{6}$ & M$a_{2}$ & 4.18 & P$a_{8}$\\
        \midrule
        \citeauthor{tanaka2019wavecyclegan2}& CycleGAN & \makecell[l]{Wave-\\ CycleGAN-\\ VC2} & \makecell[l]{L$a_{1}$,\\ L$a_{2}$,\\ L$a_{6}$} & D$a_{6}$ & M$a_{2}$ & 4.29 & \makecell[l]{P$a_{8}$,\\ P$a_{9}$}\\
        \midrule
        \multirow{4}{*}{\citeauthor{kameoka2018convs2s}} & \multirow{4}{*}{-} & \multirow{4}{*}{ConvS2S} & \multirow{4}{*}{\makecell[l]{L$a_{1}$,\\ L$a_{5}$,\\ L$a_{7}$,\\ L$a_{8}$}} & \multirow{4}{*}{D$a_{7}$} & & & \multirow{4}{*}{\makecell[l]{P$a_{10}$,\\ P$a_{11}$,\\ P$a_{12}$}}\\
        & & & & & M$a_{13}$ & 50.00 & \\
        & & & & & M$a_{14}$ & 47.00 & \\
        & & & & & & & \\
        \midrule
        \multirow{4}{*}{\citeauthor{tanaka2019atts2s}}& \multirow{4}{*}{-} & \multirow{4}{*}{AttS2S} & \multirow{4}{*}{\makecell[l]{L$a_{1}$,\\ L$a_{7}$, \\ L$a_{9}$}} & \multirow{4}{*}{D$a_{7}$} & & & \multirow{4}{*}{\makecell[l]{P$a_{10}$,\\ P$a_{11}$,\\ P$a_{12}$,\\ P$a_{13}$}} \\
        & & & & & M$a_{13}$ & 54.00 & \\
        & & & & & M$a_{14}$ & 52.00 & \\
        & & & & & & & \\
        \midrule
        \multirow{3}{*}{\citeauthor{gao2018nonparallel}} & \multirow{3}{*}{GAN} & \multirow{3}{*}{-} & \multirow{3}{*}{\makecell[l]{L$a_{1}$,\\ L$a_{2}$,\\ L$a_{5}$}} & \multirow{3}{*}{D$a_{3}$} & M$a_{2}$ & 41.25& \multirow{3}{*}{\makecell[l]{P$a_{3}$,\\ P$a_{6}$}} \\
        & & & & & M$a_{11}$ & 40.75 & \\
        & & & & & M$a_{12}$ & 47.00 & \\
        \midrule
        \citeauthor{sheng2018data} & - & - & \makecell[l]{L$a_{1}$,\\ L$a_{10}$} & D$a_{9}$ & M$a_{18}$ & 8.16 & P$a_{1}$ \\
        \midrule
        \citeauthor{sahu2018enhancing}& cGAN & - & \makecell[l]{L$a_{1}$,\\ L$a_{4}$} & D$a_{3}$ & M$a_{17}$ & 60.29 & \makecell[l]{P$a_{1}$,\\ P$a_{2}$}\\
        \midrule
        \multirow{3}{*}{\makecell[l]{Chatziagapi\\ et al.}}& \multirow{3}{*}{BAGAN} & \multirow{4}{*}{-} & \multirow{4}{*}{\makecell[l]{L$a_{1}$,\\ L$a_{2}$,\\ L$a_{6}$,\\ L$a_{11}$}} & \multirow{4}{*}{D$a_{3}$} & M$a_{2}$ & 2.88 & \multirow{4}{*}{\makecell[l]{P$a_{1}$,\\ P$a_{14}$}}\\
        & & & & & M$a_{12}$ & 66.05 & \\
        & & & & & M$a_{14}$ & 50.62& \\
        & & & & & M$a_{19}$ & 3.41 & \\
        \bottomrule
        & & & & & & & \\
\end{longtable}
\end{center}
\vspace{-20pt}
\begin{flushleft}
    \footnotesize{
    - M: Metric, RS: Results, RM:Remarks \\
    }
\end{flushleft}        

\citet{gao2018nonparallel} decomposed each speech signal into two codes: a content code that represents emotion-invariant information and a style code that represents emotion-dependent information. The content code is shared across emotion domains and should be preserved while the style code carries domain-specific information and it should change. The extracted content code of the source speech and the style code of the target domain are combined at the conversion step. Finally, they use the GAN model to enhance the quality of the generated speech.

Another widely extended research direction in speech synthesis is Voice Conversion (VC). \cite{hsu2017voice} proposed a non-parallel VC framework called Variational Autoencoding Wasserstein Generative Adversarial Network (VAWGAN). This method directly incorporates a non-parallel VC criterion into the objective function to build a speech model from unaligned data. VAWGAN improves the synthesized samples with more realistic spectral shapes. Even if VAE-based approaches can work free of parallel data and unaligned corpora, yet they have three drawbacks. First, it is difficult to learn time dependencies in the acoustic feature sequences of source and target speech. Second, the decoder of the VAEs tends to output over-smoothed results. To overcome these limitations, \cite{kameoka2018acvae} adopted fully convolutional neural networks to learn conversion rules that capture short-term and long-term dependencies. Also, by transplanting the spectral details of input speech into its converted version at the test phase, the proposed model avoids producing buzzy speech. Furthermore, in order to prevent losing class information during the conversion process, an information-theoretic regularizer is used. 

In \citeyear{kaneko2018cyclegan}, \citeauthor{kaneko2018cyclegan} made two modifications in CycleGAN architecture to make it suitable for voice conversion task and so the name CycleGAN-VC is selected for the modified architecture. Representing speech by using Recurrent Neural Networks (RNN) is more effective due to the sequential and hierarchical structure of the speech. Howsoever, RNN is computationally demanding considering parallel implementations. As a result, they used gated CNNs that are proven to be successful both in parallelization over sequential data and achieving high performance. The second modification is made by using identity loss to assure preserving linguistic information. Here, a 1D CNN is used as a generator and a 2D CNN as a discriminator to focus on 2D spectral texture. 
\begin{table*}[h]
    \setlength{\aboverulesep}{0pt}
    \setlength{\belowrulesep}{0pt}
    \caption{List of databases used for speech synthesis in the reviewed publications}
    \label{tab:sdatabases}
    \begin{tabular}{llclcl}
        \toprule
          & corpus & \#Subjects & \#samples & \#classes\\
         \toprule
         D$a_{1}$ & voice Bank (\citeauthor{veaux2013voice}) & 500 & - & 6\textsuperscript{$\bullet$} \\
         D$a_{2}$ & VC Challenge 2016 (\citeauthor{tomoki2016voice}) & 10 & 216 & - \\
         D$a_{3}$ & IEMOCAP (\citeauthor{busso2008iemocap}) & 10 & 7,142 & 9\textsuperscript{$\star$} \\
         D$a_{4}$ & FAU-AIBO (\citeauthor{schuller1988emotion}) & 51 & 18,216 & 5\textsuperscript{$\dagger$} \\
         D$a_{5}$ & VC Challenge 2018 (\citeauthor{lorenzo2018voice}) & 20 & - & - \\
         D$a_{6}$ & Japanese speech (\citeauthor{ito2017lj}) & 1 & 13,100 & - \\
         D$a_{7}$ & CMU Arctic (\citeauthor{kominek2004cmu}) & 4 & 1,132 & - \\
         D$a_{8}$ & RAVDESS (\citeauthor{livingstone2018ryerson}) & 24 & 1,440 & 8\textsuperscript{$\ddagger$} \\
         D$a_{9}$ &  Aurora 4 (\citeauthor{hirsch2000aurora}) & - & - & - \\
         \bottomrule
    \end{tabular}
    \begin{flushleft}
        \scriptsize{
        9\textsuperscript{$\star$}: anger, happiness, excitement, sadness, frustration, fear, surprise, other and neutral state\\
        5\textsuperscript{$\dagger$}: anger, emphatic, neutral, positive, rest\\
        7\textsuperscript{$\diamond$}: anger, anxiety, boredom, disgust, neutral, sadness\\
        8\textsuperscript{$\ddagger$}: neutral, calm, happy, sad, angry, fearful, surprise, and disgust
        }
    \end{flushleft}
\end{table*}

Later, they released CycleGAN-VC2 \citep{kaneko2019cyclegan} which is an improved version of CycleGAN-VC to fill the large gap between the real target and converted speech. Architecture is altered by using 2-1-2D CNN for the generator and PatchGAN for the discriminator. In addition, the objective function is improved by employing a two-step adversarial loss. It is known that downsampling and upsampling have a severe degradation effect on the original structure of the data. To alleviate this, a 2-1-2D CNN architecture is used in the generator where 2D convolution is used for downsampling and upsampling, and only 1D convolution is used for the main conversion process. Another difference is that while CycleGAN-VC uses a fully connected CNN as its discriminator, CycleGAN-VC2 uses PatchGAN. The last layer of PatchGAN employs convolution to make a patch-based decision for the realness of samples. The difference in objective functions between these two models is reported in Table \ref{tab:RawSpeechGANS}. They report that CycleGAN-VC2 outperforms its predecessor on the same database.

To overcome the shortcomings of CVAEVC\citep{kameoka2018acvae} and CycleGAN-VC \citep{kaneko2018cyclegan}, the StarGAN-VC \citep{kameoka2018stargan} method combines these two methods to address nonparallel many-to-many voice conversion. While CVAEVC and CycleGAN-VC require to know the attribute of the input speech at the test time, StarGAN does not need any such information. Other GAN-based methods for VC like WaveCycleGAN-VC \citep{tanaka2018synthetic} and WaveCycleGAN-VC2 \citep{tanaka2019wavecyclegan2} rely on learning based on filters that prevents quality degradation by overcoming the over-smoothing effect. The over-smoothing effect causes degradation in resolution of acoustic features of the generated speech signal. WaveCycleGAN-VC uses cycle-consistent adversarial networks to convert synthesized speech to natural waveform. The drawback of WaveCycleGAN-VC is aliasing distortion that is avoided in WaveCycleGAN-VC2 by adding identity loss.
\begin{table}[h]
    \setlength{\aboverulesep}{0pt}
    \setlength{\belowrulesep}{0pt}
    \caption{different losses used for speech synthesis in the reviewed publications}
    \label{tab:slosses}
    \begin{tabular}{llL{9.3cm}}
        \toprule
         & Name & Remarks\\
         \toprule
         L$a_1$ & $\mathcal{J}_{adv}$ & adversarial loss  \\
         L$a_2$ & $\mathcal{J}_{cyc}$ & cycle-consistency loss \\
         L$a_3$ & $\mathcal{J}_{cls}$ & classification loss \\
         L$a_4$ & $\mathcal{J}_{cond}$ & conditional loss \\
         L$a_5$ & $\mathcal{J}_{rec}$ & speech reconstruction loss \\
         L$a_6$ & $\mathcal{J}_{id}$ & speaker identity loss \\
         L$a_7$ & $\mathcal{J}_{att}$ & guided attention loss \\
         L$a_8$ & $\mathcal{J}_{post}$ & PostNet loss \\
         L$a_9$ & $\mathcal{J}_{s2s}$ & sequence-to-sequence loss \\
         L$a_{10}$ & $\mathcal{J}_{KL}$ & KL divergence between the output distribution and the related labels \\
         L$a_{11}$ & $\mathcal{J}_{li}$ & linguistic-information loss \\
        \bottomrule
    \end{tabular}
\end{table}

Conventional methods like VAEs, cycle-consistent GANs, and StarGAN have a common limitation. Instead of focusing on converting prosodic features like Fundamental frequency contour, they focus on the conversion of spectral features frame by frame. A fully convolutional sequence-to-sequence (seq2seq) learning approach is proposed by \cite{kameoka2018convs2s} to solve this problem. Generally, all inputs of a seq2seq model must be encoded into a fixed-length vector. In order to avoid this general limitation of seq2seq models, the authors used an attention-based mechanism that learns where to pay attention in the input sequence for each output sequence. The advantage of seq2seq models is that one can transform a sequence into another variable-length sequence. The proposed model is called ConvS2S \citep{kameoka2018convs2s} and its architecture comprises a pair of source and target encoders, a pair of source and target re-constructors, one target decoder, and a PostNet. The PostNet aims to restore the linear frequency spectrogram from its Mel-scaled version. 

Similar to ConvS2S is ATTS2S-VC \citep{tanaka2019atts2s} that employs attention and context preservation mechanisms in a Seq2Seq-based VC system. Although this method addresses the aforementioned problems, yet it has a lower performance in comparison to CVAE-VC, CycleGAN-VC, CycleGAN-VC2, and StarGAN. An ablation study is required to evaluate each component of seq2seq methods considering performance degradation.

Despite the promising performance of deep neural networks, they are highly susceptible to malicious attacks that use adversarial examples. One can develop an adversarial example through the addition of unperceived perturbation with the intention of eliciting wrong responses from the machine learning models. \citet{latif2017variational} conducted a study on how adversarial examples can be used to attack speech emotion recognition (SER) systems. They propose the first black-box adversarial attack on SER systems that directly perturbs speech utterances with small and imperceptible noises. Later, the authors perform emotion classification to clean audio utterances by removing that adversarial noise using a GAN model to show that GAN-based defense stands better against adversarial examples. Other examples of malicious attacks are simulating spoofing attacks \citep{cai2018attacking} and cloning Obama's voice using GAN-based models and low-quality data \citep{lorenzo2018can}. 
\begin{table}[h]
    \setlength{\aboverulesep}{0pt}
    \setlength{\belowrulesep}{0pt}
    \caption{List of evaluative metrics used for speech synthesis in the reviewed publications}
    \label{tab:smetrics}
    \begin{tabular}{llL{8.2cm}}
         \toprule
         & Measurement & Remarks \\
         \toprule
         M$a_{1}$ & PESQ & Perceptual Evaluation of Speech Quality\\
         M$a_{2}$ & MOS & Mean Opinion Score (MOS) for voice quality \\
         M$a_{3}$ & CSIG & MOS for prediction of the signal distortion\\
         M$a_{4}$ & CBAK & MOS for prediction of the intrusiveness of background noise\\
         M$a_{5}$ & COVL & MOS for prediction of the overall effect\\
         M$a_{6}$ & SSNR & Segmental Signal to Noise Ratio (SSNR) \\
         M$a_{7}$ & SER$_{err}$ & speech emotion classification (Error) \\
         M$a_{8}$ & SER$_{acc}$ & speech emotion classification (Accuracy) \\
         M$a_{9}$ & ID$_{err}$ & identity classification (Error) \\
         M$a_{10}$ & ID$_{acc}$ & identity classification (Accuracy) \\
         M$a_{11}$ & - & MOS for speaker similarity\\
         M$a_{12}$ & - & Preference of Emotion Conversion (\%)\\
         M$a_{13}$ & - & Preference of voice quality (\%)\\
         M$a_{14}$ & - & Preference of speaker similarity (\%)\\
         M$a_{15}$ & MCD & measures the distance between the target and converted speech\\
         M$a_{16}$ & MSD & measures the local structural differences\\
         M$a_{17}$ & UAR & Unweighted Average Recall\\
         M$a_{18}$ & WER & Weighted Error Rate\\
         M$a_{19}$ & FAD & a VGGish model to evaluate similarity metric\\
         \bottomrule
    \end{tabular}
\end{table}

The next target application of speech synthesis is data augmentation. Data augmentation is the task of increasing the amount and diversity of data to compensate for the lack of data in certain cases. Data augmentation can improve the generalization behavior of the classifiers. Despite its importance, only a few papers contributed fully toward this concept. 

\begin{table}[h]
    \setlength{\aboverulesep}{0pt}
    \setlength{\belowrulesep}{0pt}
    \caption{List of purposes and characteristics used for speech synthesis by reviewed publications }
    \label{tab:spur}
    \begin{tabular}{lL{10.55cm}}
        \toprule
        & purpose or characteristic\\
        \toprule
        P$a_{1}$ & is tested for data augmentation \\
        P$a_{2}$ & is designed for speech enhancement \\
        P$a_{3}$ & is designed for non-parallel identity preserving VC \\
        P$a_{4}$ & is designed for for defense against Malicious Adversary \\
        P$a_{5}$ & is designed for non-parallel many-to-many identity VC\\
        P$a_{6}$ & performs emotion conversion \\
        P$a_{7}$ & performance improvement \\
        P$a_{8}$ & generates vocoder-less sounding speech \\
        P$a_{9}$ & alleviates the aliasing effect \\
        P$a_{10}$ & voice conversion (VC) \\
        P$a_{11}$ & fully convolutional sequence-to-sequence \\
        P$a_{12}$ & modifies prosodic features of voice \\
        P$a_{13}$ & uses an attention-based mechanism \\
        P$a_{14}$ & generates spectrograms with high quality \\
        \bottomrule
    \end{tabular}
\end{table}

One of the researches on data augmentation for the purpose of SER improvement is the work of \cite{sheng2018data}. \citeauthor{sheng2018data} used a variant of cGANs model that works at frame level and uses two different conditions. The first condition is the acoustic state of each input frame that is combined as a one-hot vector with the noise input and fed into the generator. The same vector is combined with real noisy speech and fed into the discriminator. The second condition is the pairing of speech samples during the training process. In fact, parallel paired data is used for training. For example, original and clean speech is paired with manually added noisy speech or close-talk speech sample is paired with far-field recorded speech. The discriminator learns the naturalness of the sample based on the paired data. 

Another study with more focus on the improvement of SER is done by \citet{chatziagapi2019data}. They adopt a cGAN called Balancing GAN (BAGAN) \citep{mariani2018bagan} and improve it to generate synthetic spectrograms for the minority or under-represented emotion classes. The authors modified the architecture of BAGAN by adding two dense layers to the original generator. These layers project the input to higher dimensionality. Also, the discriminator is changed by using double strides to increase the height and width of the intermediate tensors which affect the quality of the generated spectrogram remarkably. 

Other interesting applications like cross-language emotion transfer and singing voice synthesis are also investigated by various researches. However, these applications are not thoroughly studied and they have plenty of potential for further research. One such example is ET-GAN \citep{jia2019gan}. This model uses a cycle-consistent GAN to learn language-independent emotion transfer from one emotion to another while it does not require parallel training samples. 

Also, some works are dedicated to speech synthesis in the frequency domain. Long-range dependencies are difficult to model in the time domain. For instance, MelNet model \citep{vasquez2019melnet} proves that such dependencies can be more tractably modeled in two dimensional (2D) time-frequency representations such as spectrograms. By coupling the 2D spectrogram representation and an auto-regressive probabilistic model with a multi-scale generative model, they synthesized high fidelity audio samples. This model captures local and global structures at time scales that time-domain models have yet to achieve. In a MOS comparison between MelNet and WaveNet, MelNet won with a 100\% vote for a preference of the quality of the sample.

In the case of speech synthesis in feature-domain, several pieces of research are represented under VC application. For instance, \citet{juvela2018speech} proposed generating speech from filterbank Mel-frequency cepstral coefficients (MFCC). The method starts by predicting the fundamental frequency (F$_0$) and the intonation information from MFCC using an auto-regressive model. Then, a pitch synchronous excitation model is trained on the all-pole filters obtained in turn from spectral envelope information in MFCCs. In the end, a residual GAN-based noise model is used to add a realistic high-frequency stochastic component to the modeled excitation signal. Degradation Mean Opinion Score (DMOS) is used to evaluate the quality of synthesized speech samples.

In order to evaluate the local and global structures of the generated samples, various metrics are employed by the researchers. In general, metrics like Mean opinion score (MOS) and Perceptual Evaluation of Speech Quality (PESQ) are used widely, while other efficient metrics like Mel-cepstral distortion (MCD) and modulation spectra distance (MSD) are less employed in the literature. Following, we provide a brief explanation of each metric with the hope of a more cohesive future comparison in the literature. 

MOS is a quality rating method that works based on the subjective quality evaluation test. The quality is assessed by human subjects for a given stimulus. The user rates its Quality of Experience (QoE) as a number within a categorical range with “bad" being the lowest perceived quality 1 and 5 being “Excellent" or the highest perceived quality. MOS is expressed as the arithmetic mean over QoEs.
\begin{equation}
    \text{MOS} = \frac{1}{N}\sum_{n=1}^{N}r_n,
\end{equation}
where $N$ is the total number of subjects contributed to the evaluation and $r_n$ is the QoE of the subject considering the stimuli. MOS is subject to certain biases. Number of subjects in the test and content of the samples under assessment are some of the problems. The ordinal categories code a wide range of perceptions. That is why MOS is considered to be an absolute measure of total quality, regardless of any specific quality dimension. This is useful in may applications related to communications. However, for other applications, a measure that could be more sensitive to specific quality dimensions is more suitable. Other biases include, changing the user expectation about quality over time, and the value of the smallest MOS difference that is perceptible to users and can actually claim if one method is better over another. For example, \citeauthor{pascual2017segan} and \citeauthor{macartney2018improved} achieved an MOS of 3.18 and 2.41 on the same database which provides a naive comparison of .77 MOS difference in favor of the former method. Howsoever, one question here is that whether the sample tests and the number of subjects were the same. This becomes more interesting by comparison of the methods proposed in \citet{tanaka2018synthetic} and \citet{tanaka2019wavecyclegan2} where the authors achieved only a .11 MOS difference using the same database. 

MOS is time-consuming though, it is applicable to different quantities. For instance, likewise MOS of voice quality, the MOS of signal distortion and MOS for intrusiveness of background noise are used as a metric. PESQ is an objective speech quality assessment based on subjective quality ratings. In fact, it automatically calculates what is MOS of subjective perception. PESQ integrates disturbance over several time frequency scales. This is applied by using a method that take soptimal account of the distribution of error in time and amplitude \citep{rix2001perceptual, recommendation2001perceptual}. The disturbance values are aggregated using a $L_p$ norm as follows:
\begin{equation}
    L_p = \big( \frac{1}{N} \sum_{m=1}^{N} \text{disturbance}[m]^p\big)^{\frac{1}{p}}
\end{equation}

Summing disturbance across frequency using an $L_p$
norm gives a frame-by-frame measure of perceived distortion. The subjective listening tests were designed to reduce the effect of uncertainty arising from the listener's decision by highlighting which of the three components of a noisy speech signal should form the basis of their ratings of overall quality.
\begin{table}[h]
    \setlength{\aboverulesep}{0pt}
    \setlength{\belowrulesep}{0pt}
    \centering
    \begin{tabular}{|c|l|l|l|}
    \toprule
        &
        \textbf{SGI} &
        \textbf{BAK} &
        \textbf{OVRL} \\
    \midrule
         1 & very unnatural/degraded & very conspicuous/intrusive & bad \\
    \midrule
         2 & fairly unnatural/degraded & fairly conspicuous, somewhat intrusive & poor \\
    \midrule
         3 & somewhat natural/degraded & noticeable but not intrusive & regular \\
    \midrule
         4 & fairly natural, little degradation & somewhat noticeable & good \\
    \midrule
         5 & very natural, no degradation & not noticeable & excellent \\
    \bottomrule
    \end{tabular}
    \caption{Description of SIG, BAK, and OVRL scales used in the subjective listening tests}
    \label{tab:testscales}
\end{table}
These components are the speech signal, the background noise, or both. In this method the listener successively attends and rates the synthesized speech sample on: a) the speech signal alone using a five-point scale of signal distortion (SIG), b) the background noise alone using a five-point scale of background intrusiveness (BAK), c) the overall quality using the scale of the mean opinion score (OVRL). The SIG, BAK, and OVRL scales are described in Table \ref{tab:testscales}.

The Signal to Noise Ratio (SNR) and Segmental Signal to Noise Ratio can be expressed as follows:

\begin{equation}
    \text{SNR} = 10\log_{10} \frac{\sum_{i=1}^{N}x_i^2}{\sum_{i=1}^{N}(x_i-y_i)^{2,}}
\end{equation}

where $x(i)$ and $y(i)$ are the $i$th real and synthesized samples and $N$ is the total number of samples. Segmental signal to Noise Ratio (SSNR/SegSNR) can be expressed as  the average of the SNR values of short segments (15 to 20 ms). It can be expressed as follows:

\begin{equation}
    \text{SSNR} = \frac{10}{M}\sum_{m=0}^{M-1} \log_{10} \sum_{i=Nm}^{Nm+N-1} \Bigg( \frac{\sum_{i=1}^{N}x_i^2}{\sum_{i=1}^{N}(x_i-y_i)^2} \Bigg)
\end{equation}
where $N$ and $M$ are the segment length and the number of segments, respectively. SSNR tends to provide better results than SNR for waveform encoders and generally SSNR results are poor on vocoders.

Other objective measurements include MCD that evaluates the distance between the target and converted Mel-cepstral coefficients (MCEP) sequences. Also, MSD assesses the local structural differences by calculating the root mean square error between the target and converted logarithmic modulation spectra of MCEPs averaged over all MCEP dimensions and modulation frequencies. For both metrics, smaller values indicate higher distortion between the real and converted speech. It is important to highlight that some of the successful methods like GANSynth \citep{engel2019gansynth} are not mentioned in this paper as it focuses on the musical note synthesis. 

\subsection{Audio-Visual Emotion Synthesis}
Although GAN models have an impressive performance on single-domain and cross-domain generation, yet they did not achieve much success in cross-modal generation due to the lack of a common distribution between heterogeneous data. In a cross-domain generation, one generates data samples of various styles from the same modality. As a result, the generated sample and its original counterpart have a common shape structure. However, in a cross-modal generation, the pair of samples have heterogeneous features with quite different distributions. 

In this section, we investigate the cross-modal research line where audio and video provide applications like talking heads, audio-video synchronization, facial animations, and visualizing the face of an unseen subject from their voice. Note that other modalities like text \citep{reed2016generative, gu2018look, stanton2018predicting} and biological signals \citep{palazzo2017generative} are used in combination with audio and video. However, those modalities are beyond the scope of this review paper. 

\citet{ephrat2017vid2speech} proposed the Vid2Speech model that uses neighboring video frames to generate sound features for each frame. Then, speech waveforms are synthesized from the learned speech features. In \citeyear{ephrat2017improved}, the authors designed a two-tower CNN \citep{ephrat2017improved} framework that reconstructs a natural-sounding speech signal from silent video frames of the speaking person. Their model shows that using one modality to generate samples of another modality is indeed useful because it provides the possibility of natural supervision which means segmentation of the recorded video frames and the recorded sound is not required. The two-tower CNN relies on improving the performance of a Residual neural network (ResNet) that is used as an encoder and redesigning a CNN-based decoder.

\begin{center}
    \fontsize{8}{8}\selectfont
    \begin{longtable}{L{1.8cm}L{1.5cm}L{1.75cm}L{.7cm}L{.7cm}L{.8cm}L{.85cm}L{.8cm}}
    \caption{Comparison of cross-modal Emotion synthesis models, description of loss functions (L), metrics (M), databases (D) and purposes (P) used in the reviewed publications are given in Tables \ref{tab:cmdatabases} to \ref{tab:cmpur}} \label{tab:CrossModalGANS}\\
    \toprule
        author & based on & model & loss & data & M & RS & RM\\
    \toprule \\
    \endfirsthead
    \multicolumn{8}{l}%
    {\tablename\ \thetable\ -- \textit{Continued from previous page}} \\
    \toprule
        author & based on & method & loss & data & M & RS & RM\\
    \toprule \\
    \endhead
    \hline \multicolumn{8}{c}{\textit{Continued on next page}} \\
    \endfoot
    \hline
    \endlastfoot
        \citeauthor{ephrat2017vid2speech} & - & - & L$av_{2}$ & D$av_{1}$ & M$av_{1}$ & 79.9 & P$av_{1}$ \\
        & & & & & & & \\
        \citeauthor{ephrat2017improved} & - & - & L$av_{2}$ & D$av_{1}$ & M$av_{2}$ & 1.97 & P$av_{1}$ \\
        & & & & & & & \\
        \citeauthor{wiles2018x2face} & GAN & X2Face & \makecell[l]{L$av_{2}$\\ L$av_{3}$\\ L$av_{4}$} & \makecell[l]{\\D$av_{2}$} & \makecell[l]{\\M$av_{3}$} & \makecell[l]{\\0.0521} & \makecell[l]{\\ P$av_{2}$ } \\
        \citeauthor{suwajanakorn2017synthesizing} & RNN & - & - & D$av_{1}$ & M$av_{4}$ & - & P$av_{3}$ \\
        & & & & & & & \\
        \citeauthor{vougioukas2018end} & - & - & L$av_{3}$ & D$av_{3}$ & \makecell[l]{M$av_{5}$,\\ M$av_{6}$\\ M$av_{6}$} & \makecell[l]{27.98\\ 0.844\\ $1.02e-04$ }& P$av_{1}$ \\
        & & & & & & & \\
        \citeauthor{duarte2019wav2pix} & \makecell[l]{SEAGAN\\ cGAN} & Wav2Pix & \makecell[l]{L$av_{1}$\\ L$av_{2}$} & D$av_{4}$ & M$av_{4}$ & - & P$av_{1}$ \\
        & & & & & & & \\
        \citeauthor{jamaludin2019you} & - & Vid2Speech & \makecell[l]{L$av_{1}$\\ L$av_{3}$\\ L$av_{7}$} & D$av_{2}$ & M$av_{3}$ & 327 & P$av_{4}$ \\
        & & & & & & & \\
        \bottomrule
\end{longtable}
\end{center}
\vspace{-20pt}
\begin{flushleft}
    \footnotesize{
    - M: Metric, RS: Results, RM:Remarks \\
    }
\end{flushleft}        

One of the foremost cross-modal GAN-based models is proposed by \citet{chen2017deep}. They explored the performance of cGANs by using various audio-visual encodings on generating sound/player of a musical instrument from the pose of the player or the sound of the instrument. This model is not tested on any emotional database and hence, it is not listed in Table \ref{tab:CrossModalGANS}. Another leading and interesting research work is conducted by \citet{suwajanakorn2017synthesizing}. An RNN is trained on weekly audio footage of President Barack Obama to map raw audio features to mouth shapes. In the end, a high-quality video with accurate lip synchronization is synthesized. The model can control fine-details like lip texture and mouth-pose.

Speech-driven video synthesis is the next application. The X2Face model proposed by \citeauthor{wiles2018x2face} \citeyear{wiles2018x2face} uses a facial photo or another modality sample (e.g audio) to modify the pose and expression of a given face for video/image editing. They train the model in a self-supervised fashion by receiving two samples: a source sample (video) and a driving sample (video, audio or, a combination). The generated sample inherits the same identity and style (e.g hairstyle) from the source sample and gets the pose, expression of the driving sample. The authors employed an embedding network that factorizes the face representation of the source sample and applies face frontalization. Unfortunately, the authors reported only the visual generated samples and, no further metric is used as of comparison.
\begin{table*}[h]
    \setlength{\aboverulesep}{0pt}
    \setlength{\belowrulesep}{0pt}
    \caption{List of databases used for speech synthesis in the reviewed publications}
    \label{tab:cmdatabases}
    \begin{tabular}{llclcl}
        \toprule
          & corpus & \#Subjects & \#samples & \#classes\\
         \toprule
         D$av_{1}$ & GRID (\citeauthor{cooke2006audio}) & 4 & - & - \\
         D$av_{2}$ & VoxCeleb (\citeauthor{nagrani2017voxceleb}) & 1,251 & 21,245 & - \\
         D$av_{3}$ & video footages of the President Obama & - & - & - \\
         D$av_{4}$ & Youtubers (\citeauthor{duarte2019wav2pix}) & 62 & 4,860 & - \\
         \bottomrule
    \end{tabular}
    \begin{flushleft}
        \scriptsize{
        }
    \end{flushleft}
\end{table*}

Another noteworthy work in speech-driven video synthesis is the one presented by \citet{vougioukas2018end}. They suggested an E2E temporal GAN that captures the facial dynamics and generates synchronized mouth movements and fine-detailed facial expressions, such as eyebrow raises, frowns, and blinks. The authors paired still images of a person with an audio speech to generate subject independent realistic videos. They use raw speech signal as the audio input. The model includes one generator comprising an RNN-based audio encoder, an identity image encoder, a frame decoder, and a noise generator. Also, there exist two discriminators: frame discriminator that simply classifies the frames into real and fake, and sequence discriminator distinguishing real videos from the fake ones. Evaluating the generated samples in frame-level and video-level helps to generate high-quality frames while the video remains synchronized with audio. In \citeyear{vougioukas2019realistic}, \citeauthor{vougioukas2019realistic} modified their previous work to generate speech-driven facial animations. This E2E model has the capability of generating synchronized lip movements with the speech audio and it has fine control over facial expressions like blinking and eyebrow movement.  

\begin{figure}[h] 
    \centering
    \subfigure[X2FACE]{\includegraphics[width=.9\textwidth]{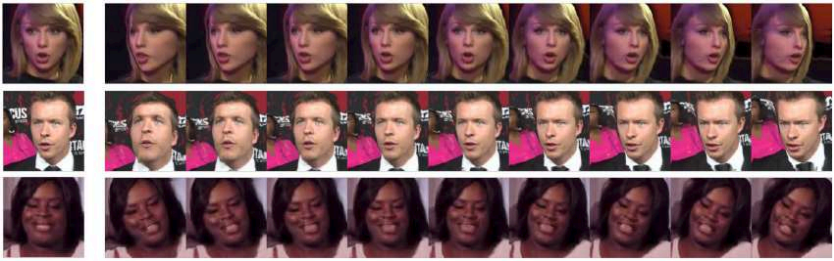}\label{subfig:x2facesample}}
    
    \subfigure[WAV2PIX]{\includegraphics[width=.4\textwidth]{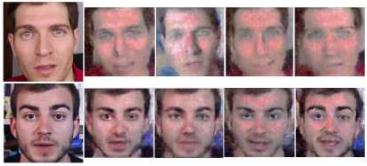}\label{subfig:wav2pixsample}} \qquad
    \subfigure[\citeauthor{vougioukas2018end}]{\includegraphics[width=.412\textwidth]{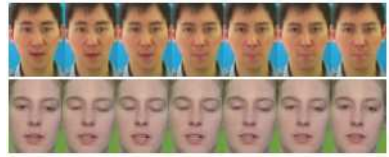}\label{subfig:e2esample}}
    
    \caption{Visual comparison of the cross-modal GAN models, images are in courtesy of the reviewed papers }\label{fig:gansamples4}
\end{figure}

\begin{table}[h]
    \setlength{\aboverulesep}{0pt}
    \setlength{\belowrulesep}{0pt}
    \caption{different losses used for cross-modal synthesis in the reviewed publications}
    \label{tab:cmlosses}
    \begin{tabular}{llL{9.3cm}}
        \toprule
         & Name & Remarks\\
         \toprule
         L$av_1$ & $\mathcal{L}_{adv}$ & adversarial loss  \\
         L$av_2$ & $\mathcal{L}_{mse}$ & mean squared error loss  \\
         L$av_3$ & $\mathcal{L}_{pix}$ & pixelwise L1 loss\\
         L$av_4$ & $\mathcal{L}_{id}$ & identity loss\\
         L$av_5$ & $\mathcal{L}_{frame}$ & frame loss\\
         L$av_6$ & $\mathcal{L}_{seq}$ & sequential loss\\
         L$av_7$ & $\mathcal{L}_{cnt}$ & content loss\\
        \bottomrule
    \end{tabular}
\end{table}
\citet{duarte2019wav2pix} proposed Wav2Pix model generating the facial image of a speaker without prior knowledge about the face identity. This is done by conditioning on the raw speech signal of that person. The model uses Least-Square GAN and SEAGAN to preserve the identity of the speaker half of the time. The generated images by this model are of low quality (See Figure \ref{fig:gansamples4}). Also, the model is sensitive to several factors like the dimensionality and quality of the training images and the duration of the speech chunk.

Likewise, \citep{jamaludin2019you} in “You said that? : Synthesising Talking Faces from Audio" designed the Speech2Vid model that gets the still images of the target face and an audio speech segment as input. The model synthesizes a video of the target face that has a synchronized lip with the speech signal. This model consists of a VGG-M as an audio encoder, a VGG-Face as an identity image encoder, and a VGG-M in reverse order as a talking face image decoder. Here instead of raw speech data, the audio encoder uses MFCC heatmap images. The network is trained with a usual adversarial loss between the generated image and the ground truth and a content representation loss.
\begin{table}[h]
    \setlength{\aboverulesep}{0pt}
    \setlength{\belowrulesep}{0pt}
    \caption{List of evaluative metrics used for cross-modal synthesis in the reviewed publications}
    \label{tab:cmmetrics}
    \begin{tabular}{llL{8.2cm}}
         \toprule
         & Measurement & Remarks \\
         \toprule
         M$av_{1}$ & SER$_{acc}$ & speech emotion classification (Accuracy) \\
         M$av_{2}$ & PESQ & automated quality evaluation\\
         M$av_{3}$ & error & L1 reconstruction error\\
         M$av_{4}$ & - & qualitative/visual representation\\
         M$av_{5}$ & PSNR & Peak Signal to Noise Ratio\\
         M$av_{6}$ & SSIM & measures image quality degradation\\
         M$av_{7}$ & ACD & content consistency of a generated video\\
         \bottomrule
    \end{tabular}
\end{table}
\begin{table}[h]
    \setlength{\aboverulesep}{0pt}
    \setlength{\belowrulesep}{0pt}
    \caption{List of purposes and characteristics used for cross-modal synthesis by reviewed publications }
    \label{tab:cmpur}
    \begin{tabular}{lL{10.55cm}}
        \toprule
        & purpose or characteristic\\
        \toprule
        P$av_{1}$ & read speech \\
        P$av_{2}$ & controlling the pose and expression of face based on audio modality\\
        P$av_{3}$ & generating high quality mouth texture\\
        P$av_{4}$ & generating unseen face of subject using raw speech\\
        P$av_{5}$ & generating lip synchronized video of a talking face \\
        P$av_{6}$ & generating an intelligible speech signal from silent video of a speaking person \\
        \bottomrule
    \end{tabular}
\end{table}

One of the most important models developed in the cross-modal community is the SyncGAN model \citep{chen2018syncgan} capable of successfully generating synchronous data. A common problem of the aforementioned cross-modal GAN models is that they are one-directional because they learn the transfer between different modalities. This means they cannot generate a pair of synchronous data from both modalities simultaneously. SyncGAN addresses this problem by learning in a bidirectional mode and from synchronous latent space representing the cross-modal data. In addition to the general generator and discriminator of the vanilla GAN, the model uses a synchronizer network for estimating the probability that two input data are from the same concept. This network is trained using synchronous and asynchronous data samples to maximize the following loss function:

\begin{align}
    \mathcal{L}_{S} =& \mathbb{E}_{\x_1 \sim p_{r}(\x_1),\x_2 \sim p_{r}(\x_2)}[\log S(\x_1^i, \x_2^j)\mid i=j] \nonumber \\ + &
    \mathbb{E}_{\x_1 \sim p_{r}(\x_1),\x_2 \sim p_{r}(\x_2)}[\log(1-S(\x_1^i, \x_2^j)) \mid i\neq j]
\end{align}

Similarly, CMCGAN \citep{hao2018cmcgan} is a cross-modal CycleGAN that handles generating mutual generation of cross-modal audio-visual videos. Given an image/sound sample from a musical instrument outputs a sound LMS or an image of the player. Unfortunately, neither SyncGAN nor CMCGAN are tested on any multi-modal emotional database. 

In Figure \ref{fig:gansamples4} we compared the generated samples of the reviewed publications qualitatively.

\section{Discussion}\label{sec:disc}
\begin{figure}
    \centering
    \includegraphics[width=.95\textwidth]{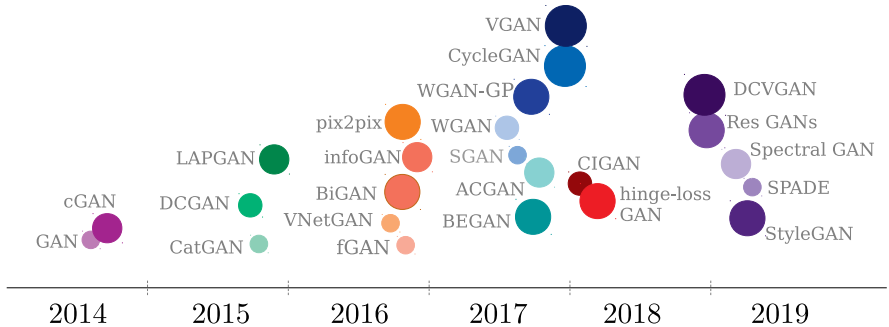}
    \caption{Evolution of GAN models, the horizontal line shows the release year of the model, each specific year is shown using shades of one color}
    \label{fig:ganevolution}
\end{figure}

In this section, we discuss the concepts that are yet not explored thoroughly about GAN-based emotion synthesis within the literature. Also, despite the active development of GAN models, there exist open research problems like mode collapse, convergence failure, and vanishing gradients. Following we discuss these problems. Also, the evolution of GAN models is shown in Figure \ref{fig:ganevolution}. 

\subsection{Disadvantages}\label{ssec:discdis}
The most important drawback of GAN models is mode drop or mode collapse. Mode collapse occurs when a generator learns to generate a limited variety of samples out of the many modes available in the training data. \citet{roth2017stabilizing} attempted to solve the mode collapse problem by stabilizing the training procedure using regularization. Numerical analysis of general algorithms for training GAN showed that not all training methods actually converge \citep{mescheder2018training} which leads to mode collapse problem. Several objective functions \citep{berthelot2017began} and structures \citep{ghosh2018multi} are developed to tackle this problem, however, none have solved the problem thoroughly.

GANs also suffer from convergence failure. Convergence failure happens when the model parameters oscillate and they cannot stabilize during training. In the minimax game, convergence occurs when the discriminator and the generator reach the optimal point under Nash equilibrium theorem. Nash equilibrium is defined as the situation where one player will not change action irrespective of opponent action.

It is known that if the discriminator performs too accurately, the generator fails due to the vanishing gradient. In fact, the discriminator does not leak/provide enough information for the generator to continue the learning process. 

\subsection{Open Research Problems}
In addition to the theoretical problems mentioned in section \ref{ssec:discdis}, GANs have task-based limitations. For instance, GANs cannot synthesize discrete data like one-hot coded vectors. Although this problem is addressed partially in some research works \citep{kusner2016gans, jang2016categorical,maddison2016concrete}, yet it needs more attention to unlock the full potential of GAN models. A series of novel divergence algorithms like FisherGAN \citep{mroueh2017fisher} and the model proposed by \citet{mroueh2017sobolev} try to improve the convergence for training GANs. This area deserves more exploration by studying families of integral probability metrics.

The objective of a GAN model is to  generate new samples that come from the same distribution as the training data. However, they do not generate the distribution that generated the training examples. As a result, they don't have a prior likelihood or a well-defined posterior. The question here is how can one estimate the uncertainty of a well-trained generator.

Considering the emotion synthesis domain, some problems are studied partially. First of all, data augmentation is not fully explored. At the time of writing this paper, there is no large scale image database generated artificially by using GAN models and released for public usage. Such a database could be compared in terms of classification accuracy and quality with the existing databases. Although methods like GANimation and StarGAN successfully generate all sorts of facial expressions, yet generating a fully labeled database requires further processing. For example, the synthesized samples should be annotated and tested against a ground truth like facial Action Units (AU) to confirm that the generated samples carry the predefined standards of a specific emotional class. This issue becomes very complicated when one deals with compound emotions and not only the basic discrete emotions. Also, generated samples are not evaluated within continuous space considering the arousal, valence, and dominance properties of the emotional state. Finally, despite the fact that some successful GAN models are proposed for video generation, the results are not realistic.

In the case of speech emotion synthesizing, majority of papers focused on raw speech and spectrgorams. As a result, feature-based synthesis is less explored. Human-likeliness of the generated speech samples is another open discussion in this research direction. Furthermore, evaluation metrics in this field is less developed and merely the ones from the traditional speech processing are used on the generated results. Research works that are focusing on cross-modal emotion generation do not exceed from few publications. This research direction requires both developing new ideas and improving the result of previous models.

\subsection{Applications}\label{ssec:discapp}
One important application of GAN models in the computer vision society includes synthesizing Super Resolution (SR) or photo-realistic images. For example, SRGAN \citep{ledig2017photo} and Enhanced SRGAN \citep{wang2018esrgan} are generating photo-realistic natural images for an upscaling factor. Considering the facial synthesis, these applications include manipulation of facial pose using DRGAN \citep{tran2017disentangled} and TPGAN \citep{huang2017beyond}, generating a facial portrait \citep{yi2019apdrawinggan}, generating face of an artificial subject or manipulating the facial attributes of a specific subject \cite{radford2015unsupervised}, \citep{choi2018stargan}, and synthesizing/manipulating fine detail facial features like skin, lip or teeth texture \citep{suwajanakorn2017synthesizing}. Generally speaking, the application of GAN considering the visual modality could be categorized to texture synthesis, image super-resolution, image inpainting, face aging, face frontalization, human image synthesis, image-to-image translation, text-to-image, sketch-to-image, image editing, and video generation. Some specific applications with respect to the emotional video generation include the synthesizing of talking heads \citep{tulyakov2018mocogan}, \citep{pumarola2018ganimation}.

In the case of speech emotion synthesis, as mentioned before these applications can be categorize to speech enhancement, data augmentation, and voice conversion. Other research directions like feature learning, imitation learning, and reinforcement learning are important research directions for the near future.

\section{Conclusion}
In this paper, we survey the state-of-the-art proposed in human emotion synthesis using GAN models. GAN models proposed first in \citeyear{goodfellow2014generative} by \citeauthor{goodfellow2014generative}. The core idea of GANs is based on a zero-sum game in game theory. Generally, a GAN model consists of a generator and a discriminator, which are trained iteratively in an adversarial learning manner, approaching Nash equilibrium. Instead of estimating the distribution of real data samples, GANs learn to synthesize samples that adapt to the distribution of real data samples. Fields like computer vision, speech processing, and natural language processing benefit from the ability of GAN in generating infinite new samples from potential distributions.

\begin{acknowledgements}
We would like to thank Eastern Mediterranean University for supporting this research work through the BAP-C project under the grant number BAP-C-02-18-0001. This work was partially supported by Fundação de Amparo à Pesquisa do Estado de São Paulo (FAPESP) under Grants 2018/26455-8 and 2020/01928-0.
\end{acknowledgements}

%
%

\scriptsize{

\bibliographystyle{spbasic}      
\bibliography{template}   
}

\end{document}